\documentclass[journal]{vgtc}                     

\onlineid{1174}

\def\etal{\emph{et al.}\xspace}

\vgtccategory{Research}

\vgtcpapertype{algorithm/technique}

\title{NIS-SLAM: Neural Implicit Semantic RGB-D SLAM for 3D Consistent Scene Understanding}

\author{\authororcid{Hongjia Zhai}{0000-0002-7729-8787}, \authororcid{Gan Huang}{0000-0001-8515-2721}, \authororcid{Qirui Hu}{0009-0009-1854-689X}, \authororcid{Guanglin Li}{0009-0000-8996-3775}, \authororcid{Hujun Bao}{0000-0002-2662-0334}, and \authororcid{Guofeng Zhang}{0000-0001-5661-8430}
}
\authorfooter{
  \item  Hongjia Zhai, Gan Huang, Qirui Hu, Guanglin Li, Hujun Bao, and Guofeng Zhang are with the State Key Lab of CAD\&CG, Zhejiang University, Hangzhou 310058, China. E-mails: \{zhj1999, huanggan, 3200105912, liguanglin, baohujun, zhangguofeng\}@zju.edu.cn.
  
  \item  Corresponding Author: Guofeng Zhang.
}

\abstract{In recent years, the paradigm of neural implicit representations has gained substantial attention in the field of Simultaneous Localization and Mapping (SLAM). However, a notable gap exists in the existing approaches when it comes to scene understanding. In this paper, we introduce NIS-SLAM, an efficient neural implicit semantic RGB-D SLAM system, that leverages a pre-trained 2D segmentation network to learn consistent semantic representations. Specifically, for high-fidelity surface reconstruction and spatial consistent scene understanding, we combine high-frequency multi-resolution tetrahedron-based features and low-frequency positional encoding as the implicit scene representations. Besides, to address the inconsistency of 2D segmentation results from multiple views, we propose a fusion strategy that integrates the semantic probabilities from previous non-keyframes into keyframes to achieve consistent semantic learning. Furthermore, we implement a confidence-based pixel sampling and progressive optimization weight function for robust camera tracking. Extensive experimental results on various datasets show the better or more competitive performance of our system when compared to other existing neural dense implicit RGB-D SLAM approaches. Finally, we also show that our approach can be used in augmented reality applications. Project page: \href{https://zju3dv.github.io/nis_slam}{https://zju3dv.github.io/nis\_slam}.
}


\teaser{
  \centering
  \includegraphics[width=\linewidth]{teaser.pdf}
  \caption{We present NIS-SLAM, a neural implicit semantic RGB-D SLAM system that incrementally reconstructs the environment with 3D consistent scene understanding.
  As shown in the figure, taking continuous RGB-D frames and 2D noise segmentation results as input, our system can reconstruct high-fidelity surface and geometry, learn 3D consistent semantic field, and recover the objects in the scene.
  }
  \label{fig:teaser}
}

\renewcommand{\manuscriptnotetxt}{}

\graphicspath{{figs/}{my_figs}{figures/}{pictures/}{images/}{./}} 

\usepackage{tabu}                      
\usepackage{booktabs}                  
\usepackage{lipsum}                    
\usepackage{mwe}                       

\usepackage{graphicx}
\usepackage{subfigure} 
\usepackage{amsmath,amsfonts}
\usepackage{multirow} 
\usepackage{makecell} 
\usepackage[table,svgnames]{xcolor} 
\usepackage{colortbl} 
\usepackage{diagbox} 
\usepackage{arydshln} 
\usepackage{tabularx}
\usepackage{booktabs} 

\colorlet{colorFst}{LightGreen!35}       
\colorlet{colorSnd}{LightBlue!75} 
\colorlet{colorTrd}{LightGrey!65}      
\colorlet{colorLow}{darkgray!30}    
\newcommand{\fs}{\cellcolor{colorFst}\bf}   
\newcommand{\nd}{\cellcolor{colorSnd}}      
\newcommand{\rd}{\cellcolor{colorTrd}}      

\usepackage{mathptmx}                  

\begin{document}


\firstsection{Introduction}

\maketitle
Dense Simultaneous Localization and Mapping (SLAM) is a fundamental research topic in 3D computer vision, which aims to localize the 6 degrees of freedom (6DoF) pose of the camera and reconstruct a dense map in an unknown environment.
It is an essential part of Virtual/Augmented Reality (VR/AR), robot localization/navigation, and visual perception.
For example, in VR/AR, the SLAM system can provide pose information for mobile devices and head-mounted devices to help users interact with virtual content, and the geometric results of dense reconstruction can better simulate physical space collisions, thereby creating a better immersive and engaging experience.

Traditional RGB-D SLAM~\cite{dai:2017:bundlefusion,newcombe:2011:KinectFusion,whelan2012kintinuous} usually performs frame-to-model optimization to track camera based on the pixel-level depth or color observations from RGB-D camera.
Benefit from the depth sensor, traditional approaches use iterative closest points (ICP)~\cite{icp} and truncated signed distance function (TSDF)~\cite{curless1996volumetric} to update a global map represented by geometric primitives, such as voxels, cost volumes, and surfels.
With the development of deep learning, recent works have turned to exploring data-driven priors and leveraging the smoothness properties of neural networks~\cite{huang2021di-fusion,bloesch:2018:codeslam,teed:2021:droid,Weder2021NeuralFusion,bdloc}. 
Although traditional and learning-based dense SLAM systems have shown good localization and reconstruction performance, they cannot perform novel view rendering and produce watertight surfaces.
For VR/AR applications, realistic image rendering and novel view synthesis also play an important role. 


As one of the research fields, Neural Radiance Field (NeRF) based SLAM methods~\cite{sucar:2021:imap,co-slam,eslam,point-slam} have shown significant advantages in novel view rendering, high-fidelity map reconstruction, and hole filling.
Different from the traditional SLAM approaches, neural implicit SLAM uses a neural radiance field to represent scene property and uses multilayer-perceptron (MLP) to decode the attributes of the scene (color, density, signed distance function, semantic information, etc.). 
Benefiting from volume rendering~\cite{volume_rendering} and positional encoding~\cite{mildenhall:2020:nerf}, implicit representation enables high-fidelity image rendering and new perspective synthesis.
iMAP~\cite{sucar:2021:imap} is the first neural implicit dense SLAM, which directly uses a single MLP to model the geometry and appearance information.
However, it is hard to reconstruct large indoor scenes due to the forgetting problem of MLP in continual learning.
To overcome this challenge, many works propose to use additional parametric encoding, like dense feature grid/plane/point~\cite{zhu:2021:niceslam,eslam,point-slam,vox-fusion,nicer-slam}, hash table~\cite{co-slam,h2mapping,neural-sem-slam}, to increase the representation ability of the MLP.
However, additional dense features will lead to more memory usage, which is not efficient.
Besides, semantic information also plays an important role in SLAM, which allows robots to perceive and understand the world.
Although there exist works~\cite{zhi2021semantic-nerf,siddiqui2023panoptic-lifting,fu2022panopticnerf,kundu2022panopticneuralfiled} that combine neural implicit representation and semantics modeling, little attention has been paid to modeling semantic information in a neural implicit dense SLAM.
This is because using neural implicit representation to simultaneously perform camera tracking and semantic learning with inconsistent 2D semantic input is rather challenging.

vMAP~\cite{kong2023vmap} and Haghighi \etal ~\cite{neural-sem-slam} are the most related works that use implicit representation to model 3D instance/semantic information.
vMAP~\cite{kong2023vmap} uses ground truth pose and instance mask for commonly used datasets~\cite{julian:2019:replica, dai:2017:scannet} to reconstruct the object in the scene.
Haghighi~\cite{neural-sem-slam} uses ORB-SLAM3~\cite{orb-slam3} to provide pose for the mapping process and they ignore the inconsistency problem of 2D segmentation.
Using the traditional SLAM approach as a decoupled front end makes them turn into a mapping method instead of an SLAM method.
Besides, they don't handle the problem of camera pose estimation and multi-view inconsistency of 2D segmentation.
As mentioned above, currently, there is no implicit-based approach that simultaneously performs camera tracking, dense surface reconstruction, and 3D consistent scene semantic understanding from noisy 2D input.

To this end, in this paper, we introduce NIS-SLAM, an efficient neural implicit semantic RGB-D SLAM system, which enables simultaneous scene reconstruction and learning 3D consistent semantics from the inconsistent segmentation results of 2D CNN.
Specifically, for high-fidelity surface reconstruction and spatial consistent scene understanding, we use the hybrid representation of high-frequency multi-resolution tetrahedron-based features and low-frequency positional encoding as the input of our system.
Besides, to learn a 3D consistent semantic field, an effective multi-view semantic fusion strategy is introduced to handle the inconsistency of 2D segmentation results.
Finally, semantic-guided pixel sampling and progressive optimization weight are also used for robust camera tracking.
In summary, our contributions include:
\begin{itemize}
  \item We propose an efficient neural implicit semantic RGB-D SLAM system with hybrid implicit representation. Our system can simultaneously reconstruct the environment and model 3D consistent semantic information from 2D segmentation results.
  \item We propose an effective multi-view semantic fusion that enables learning 3D consistent semantic information. Besides, semantic guided sampling and progressive optimization weight are used for robust camera tracking.
  \item We perform extensive experiments on commonly used datasets to demonstrate the state-of-the-art and comparable performance of camera tracking, reconstruction, and semantic segmentation.
\end{itemize}

This paper is structured as follows: In~\cref{sec:related_work}, we provide a review of related works to contextualize our research.
Next, in \cref{sec:method}, we explain each key component in our reconstruction pipeline.
Subsequently, in \cref{sec:exp}, we evaluate the performance of our system through various synthetic and real-world datasets.
Finally, conclusion is given in \cref{sec:conclusion}.
\section{Related Work}
\label{sec:related_work}

\subsection{Dense Visual SLAM}
DTAM~\cite{dtam} was the first dense SLAM system that leveraged pixel-level photometric information to perform camera pose tracking. 
Taking advantage of RGB-D cameras, some traditional approaches use iterative closest points (ICP)~\cite{icp} and truncated signed distance function (TSDF)~\cite{curless1996volumetric} to achieve good 3D reconstruction results, such as KinectFusion~\cite{newcombe:2011:KinectFusion}, BundleFusion~\cite{dai:2017:bundlefusion}, and ElasticFusion~\cite{whelan:2016:ElasticFusion}.
Benefiting from deep learning technology, some learning-based approaches~\cite{tateno2017cnnslam,Weder2021NeuralFusion,sun2021NeuralRecon,teed:2021:droid,huang2021di-fusion,bloesch:2018:codeslam} are proposed to improve the robustness and accuracy. 
For example, DI-Fusion~\cite{huang2021di-fusion} and CodeSLAM~\cite{bloesch:2018:codeslam} learn the scene representation implicitly which encodes the geometry information in a low dimensional latent feature.
BAD-SLAM~\cite{schops:2019:badslam} uses a direct bundle adjustment (BA) to jointly optimize the poses of keyframe and 3D scene geometry. 
There are still some approaches that pay attention to scene understanding, like SemanticFusion~\cite{mccormac2017semanticfusion} and PanopticFusion~\cite{narita2019panopticfusion}. 
Besides, semantic-based object alignment~\cite{han2021reconstructingpanopticmapping} and reconstruction~\cite{grinvald2019volumetric} also play an important role in robot navigation and perception.
Although these methods use learning-based components, the scene and map representation of the environment and the system framework are still the same as traditional SLAM methods. 
They cannot achieve realistic image rendering and novel view synthesis, which is more important for the applications of VR/AR.

\subsection{Neural Implicit Scene Representation}
Implicit representation has facilitated novel applications in 3D computer vision, such as novel view synthesis~\cite{mildenhall:2020:nerf,turki:2022:meganerf,tancik:2022:blocknerf}, surface reconstruction~\cite{azinovic:2022:neuralrgbd,oechsle:2021:unisurf,wang:2021:neus,Vox-Surf,chen2024pgsr}, and dynamic scene reconstruction~\cite{park2021hypernerf,gao2021dynamic,pumarola2021dnerf,park2021nerfies}.
Depending on the scene representation approaches, three main categories emerge: network-based, grid-based, and point-based. 
Network-based methods~\cite{mildenhall:2020:nerf,martin2021nerfw,park2021hypernerf,wang:2021:neus,azinovic:2022:neuralrgbd} offer a continuous representation of the scene through implicit modeling via coordinate MLPs, enabling rendering and surface reconstruction. 
However, due to the limited representation capacity of MLPs, they often exhibit reduced efficacy in representing and reconstructing large-scale scenes.
Consequently, grid-based approaches have evolved, employing explicit representations like dense grids to encode scene information, partially mitigating the limitations of MLP-based decoding.
Additionally, some methods utilize sparse voxels~\cite{liu:2020:nsvf,Vox-Surf,imtooth}, hashtables~\cite{mueller2022instantngp,rosu2023permutosdf}, tetrahedron~\cite{tetra-nerf,nvdiffrec,dmtnet} or other efficient representations~\cite{tensorf,eslam} to reduce the parameters of the grid.
Point-based methods~\cite{xu2022pointnerf,point-slam} allow for a more flexible scene representation, where points can be concentrated near complex surfaces to avoid wasting memory on modeling unused free space.
3D Gaussian Splatting~\cite{kerbl3Dgaussians} also can be viewed as the point-based method, which uses Gaussian to encode the local information of 3D points in the scene.
The splatting strategy avoids dense point sampling along the camera ray, which accelerates the rendering speed of the neural implicit model.
However, the overall parameter count for the entire representation is directly proportional to the number of points which usually needs to be at least semi-dense to ensure a relatively good reconstruction performance.

\begin{figure*}[ht!]
\centering
 \includegraphics[width=\linewidth]{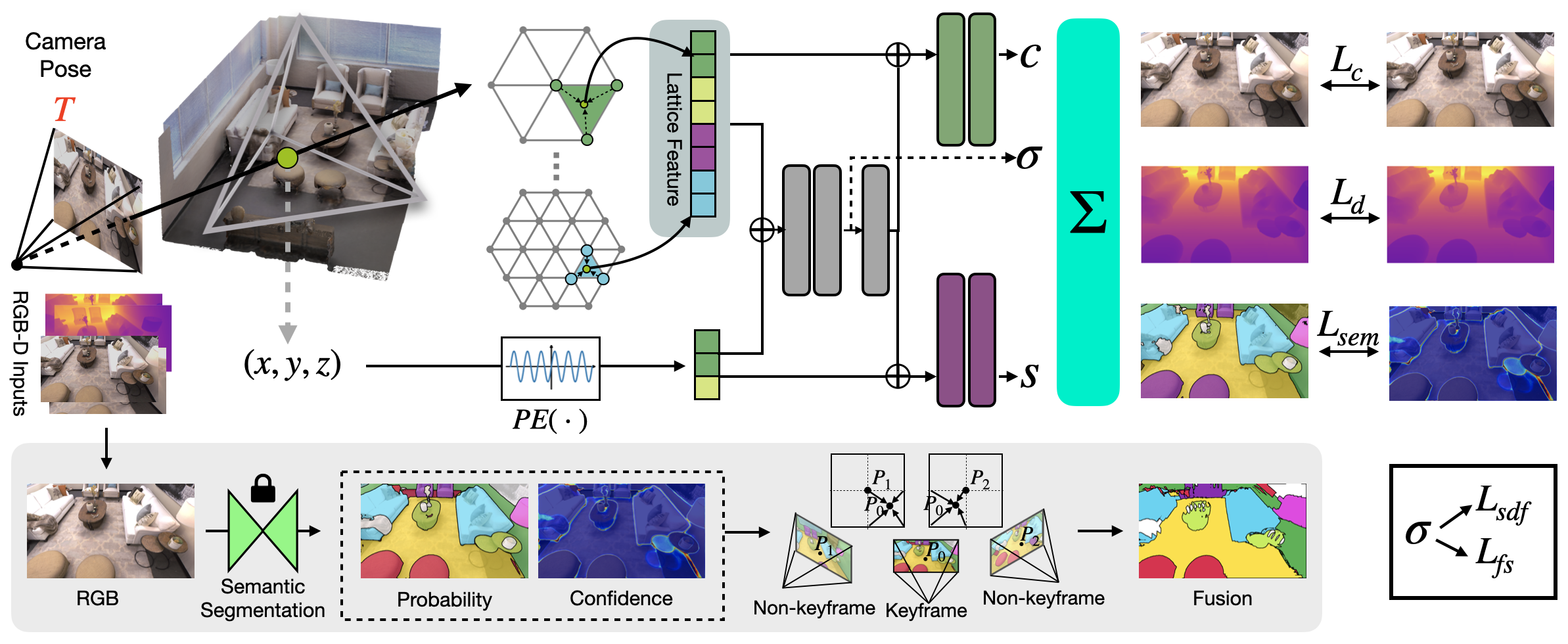}\\
\caption{Our system takes RGB-D frames as input to perform camera tracking and mapping via volume rendering and models 3D semantics with the noise 2D segmentation results from Mask2Former~\cite{cheng2021mask2former}. 
Based on the hybrid implicit representation of multi-resolution tetrahedron feature $\theta$ and positional encoding $\texttt{PE}(p)$, we decode the SDF $\sigma$, latent feature $h$, color $c$, and semantic probability $s$ with three MLPs \{$\mathcal{M}_{geo}$, $\mathcal{M}_{color}$, and $\mathcal{M}_{sem}$\}.
To model consistent semantic property, we fuse multi-view semantics of nearby non-keyframes for learning 3D consistent representation.
}
\label{fig:pipeline}
\end{figure*}

\subsection{Neural Implicit Dense SLAM}
Recently, there have been successful approaches that combine dense visual SLAM with neural implicit radiance fields~\cite{sucar:2021:imap,vox-fusion,co-slam,eslam}.
Compared to the traditional SLAM approaches, neural implicit SLAM can reconstruct the watertight surface of the scene and perform realistic novel view rendering.
iMAP~\cite{sucar:2021:imap} was the first neural implicit SLAM approach that uses coordinate MLPs to represent the scene and continuously update it during incremental mapping. 
To address the challenge of reconstructing large indoor scenes, some approaches utilize multi-level dense feature grids~\cite{zhu:2021:niceslam,nicer-slam,eslam}, hash table~\cite{co-slam}, and voxel/point embeddings~\cite{vox-fusion,h2mapping,point-slam,vox-fusion++} to mitigate the representation ability of coordinate MLPs.
Besides, some methods~\cite{h2mapping,Orbeez-slam,nerf-slam,zhang2023go-slam,kong2023vmap} decouple the front and back ends, employing traditional or learning-based VO/VIO~\cite{orb-slam2,shan2019vins-rgbd,teed:2021:droid} for camera pose tracking and utilizing implicit representation in the back end for dense reconstruction. 
To perform scene understanding, Haghighi \etal~\cite{neural-sem-slam} and vMAP~\cite{kong2023vmap} incorporate semantic/instance information into neural implicit reconstruction, which both use ORB-SLAM3~\cite{orb-slam3} to provide camera pose.
Zhu~\etal~\cite{zhu2024sni} propose the first neural semantic mapping system, which learns consistent semantic fields based on the pose estimated by neural SLAM.
These kinds of approaches using decoupled modules can accelerate tracking speed and accuracy in complex scenes but depart from a couple of pure implicit SLAMs.

\section{Method}
\label{sec:method}

The pipeline of our method is shown in ~\cref{fig:pipeline}.
Given an input sequence of RGB-D images \{$I_i \in \mathbb{R}^3, D_i \in \mathbb{R}$\}, we first generate the 2D semantic segmentation results \{$S_i$\} and confidence \{$Conf_i$\}  via a pre-trained CNN model (Mask2Former~\cite{cheng2021mask2former}).
Based on those inputs, our system is capable of recovering camera poses, and implicit SDF fields, and additionally reconstructing a 3D consistent semantic representation.
These individual components will be explained in detail in the following subsections.
We first introduce our tetrahedron-based neural implicit representation (\cref{subsec:representation}) and how color/depth/semantic information is rendered via SDF-based volume rendering (\cref{subsec:volume_rendering}).
Then, the multi-view semantic fusion strategy is proposed to handle the noise semantic information (\cref{subsec:sem_fusion}).
Besides, the overall optimization objective functions are introduced in~\cref{subsec:objective_function}.
Finally, the tracking and mapping processes of our system are shown in \cref{subsec:tracking} and \cref{subsec:mapping}, respectively.


\begin{figure}[ht!]
\centering
\includegraphics[width=0.9\linewidth]{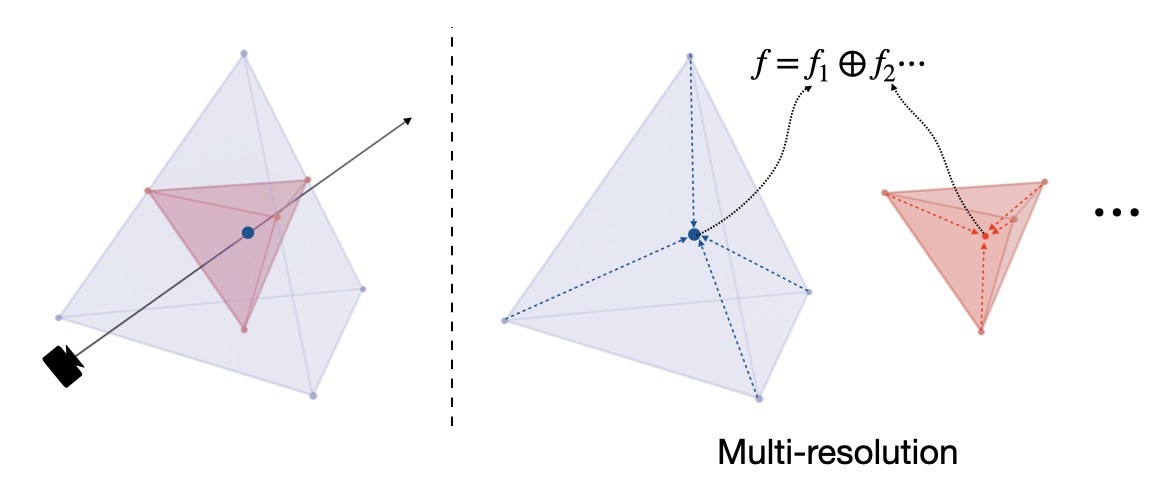}
\caption{Barycentric interpolation of multi-resolution tetrahedron feature. 
For tetrahedrons of different resolutions, we obtain the features of the query point according to the vertex features and barycentric.}
\label{fig:interpolation}
\end{figure}
\subsection{Tetrahedron-based Representation}
\label{subsec:representation}
Only using MLPs~\cite{sucar:2021:imap} to decode the scene property can lead to worse performance due to the forgetting problem during incremental reconstruction of large scenes.
Although grid-based or point-based methods~\cite{co-slam,vox-fusion,point-slam} can enhance the encoding capacity of MLPs for large indoor scenes, they can not fit the scene surface very well due to the inherent flaws of their representation.
Using tetrahedron~\cite{tetra-nerf,dmtnet} to encode the scene information is more appropriate than using cube/grid/point due to the fundamental geometric primitive is the triangular face.

Same to ~\cite{mueller2022instantngp,rosu2023permutosdf}, we employ a multi-resolution feature to encode high-frequency information (\textit{e.g.} color) and use a hash table to reduce the additional parameters.
Besides, benefiting from the joint encoding~\cite{co-slam}, we additionally apply a low-frequency positional encoding~\cite{mildenhall:2020:nerf} for reasoning spatial semantics.
Additionally, as shown in~\cite{bakefuture}, the semantic information is the low-frequency information compared to the color.
So, for semantic property decoding, we concatenate the positional encoding with the latent feature as the input. 
For color property decoding, we concatenate the tetrahedron feature with the latent feature as the input. 
Those processes are shown in the top part of~\cref{fig:pipeline}.

Specifically, for a 3D point $p\in\mathbb{R}^3$, we first compute its low-frequency positional encoding value $\texttt{PE}(p)$ with following equation:
\begin{equation}
    \texttt{PE}(p) = [\gamma(x), \gamma(y), \gamma(z)],
\end{equation}
\begin{equation}
    \gamma(x) = (\sin (2^0\pi x), \cos ((2^0\pi x)), \cdots, \sin (2^{n-1}\pi x), \cos ((2^{n-1}\pi x))),
\end{equation}
where $n$ represents the frequency, and we set $n$=8 in our experiments.

Then, we interpolate the high-frequency feature of $p$ from multi-resolution tetrahedron features $\Theta$ = \{$\theta_l$\}$_{l=1}^L$.
Here, we retrieve multi-resolution tetrahedrons that contain the point $p$.
We obtain its feature via barycentric interpolation according to the feature in the vertexes of the tetrahedrons
with the following equation $\mathcal{F}(p;\Theta)$.
\begin{equation}
    \mathcal{F}(p;\Theta) = [\mathcal{V}(p,\theta_1), \cdots, \mathcal{V}(p,\theta_l)],
\end{equation}
where $\mathcal{V}(\cdot)$ refers to the barycentric interpolation function for each resolution tetrahedron.
The barycentric interpolation of the tetrahedron is shown in~\cref{fig:interpolation}.
After obtaining the multi-resolution features, we decode the geometry property with geometry MLP $\mathcal{M}_{geo}$:

\begin{equation}
    (h,\sigma) = \mathcal{M}_{geo}(\texttt{PE}(p), \mathcal{F}(p;\Theta)),
\end{equation}
where $h$ and $\sigma$ are the latent feature vector and signed distance of location $p$.
Finally, we decode the color and semantic probability based on the latent feature $h$ and input encodings via separated MLPs:
\begin{equation}
    c = \mathcal{M}_{color}(\mathcal{F}(p;\Theta), h), \quad  s=\mathcal{M}_{sem}(\texttt{PE}(p), h).
\end{equation}

So, the optimizable parameters for the whole implicit representation are three MLP decoders \{$\mathcal{M}_{geo}$, $\mathcal{M}_{color}$, $\mathcal{M}_{sem}$\} and multi-resolution tetrahedron features $\Theta$.

\subsection{SDF-based Volume Rendering}
\label{subsec:volume_rendering}
Compared to occupancy or density-based rendering approaches~\cite{zhu:2021:niceslam,point-slam}, using SDF-based volume rendering can lead to more accurate surface reconstruction.
To render color, depth, and semantics for each pixel, we adopt the SDF-based volume rendering strategy~\cite{azinovic:2022:neuralrgbd,wang:2021:neus} to aggregate the property of points along the ray.

Specifically, for ray $r$ emitted from the center of camera with origin $o_r\in \mathbb{R}^3$ and view direction $d_r \in \mathbb{R}^3$, we can sample points along $r$:
\begin{equation}
    p_r(i) = o_r + z_r(i)\cdot d_r, \quad i\in\{1, \cdots, M\},
\end{equation}
where $M$ is the number of total sampled points, $z_r(i) \in \mathbb{R}$ is the depth of $i$-th sampled point.
Similar to other approaches~\cite{co-slam,eslam,zhu:2021:niceslam}, we replace the importance sampling with a depth-guided sampling strategy.
Specifically, we first uniformly sample $M_c$ points between the \textit{near} and \textit{far} bound. 
We then uniformly sample $M_f$ points around the surface according to the depth values of the selected pixel.

For all $M= M_{c}+M_{f}$ points along the ray, we query their signed distance, color, and semantic probability values using MLP decoders (\cref{subsec:representation}).
As shown in~\cite{azinovic:2022:neuralrgbd}, a simple bell-shaped weight function can be employed to transform the signed distance function, $\sigma$, into rendering weights.
Here, we adopt the Gaussian distribution with mean 0 as the rendering weight distribution, which is shown as follows:
\begin{equation}
   \widetilde{w}_i=\frac{w_i}{\sum_j w_j} \quad \text{with} \quad w_i = \frac{1}{\delta \sqrt{2\pi}}\exp(-\sigma_i^2/\delta^2),
\end{equation}
where $\delta$ is the parameter, which controls the shape of the weight distribution for points with different signed distance values.
$\widetilde{w}_i$ is the normalized rendering weight for $i$-th point.

So, for ray $r$, we can obtain its scene property via:
\begin{equation}
    \hat{c}= \sum_{i=1}^{M} \widetilde{w}_{i} c_{i}, \quad \hat{d}= \sum_{i=1}^{M} \widetilde{w}_{i} d_{i}, \quad \hat{s} = \sum_{i=1}^{M} \widetilde{w}_{i} s_{i}.
    \label{eq:render_propetry}
\end{equation}

\subsection{Multi-view Noisy Semantic Fusion}
\label{subsec:sem_fusion}
Due to the inherent noise and multi-frame inconsistencies of results generated by Mask2Former~\cite{cheng2021mask2former}, the SLAM system may learn wrong 3D semantic representations when the segmentation results of keyframes are inaccurate.
Inspired by the warp strategy~\cite{zhou2017sfmlearner,wei2020ffwm}, we propose an efficient approach to fuse semantic information from multi-view frames to address the inconsistency issue.
At the bottom of~\cref{fig:pipeline}, we first back project the pixel, $P_t$, of keyframe $I_t$ into $k$ previous non-keyframes $I_{t-i}$ ($i=\{1,...,k\}$) according to the relative pose, $T_{t\rightarrow t-i}$, estimated by our neural implicit SLAM system.
Then, we synthesize the projected semantics via the differentiable bilinear sampling proposed in~\cite{jaderberg2015stn}. 
The formula is shown in the following: 
\begin{equation}
S_{t-i \rightarrow t} = \texttt{interp}(KT_{t\rightarrow t-i}D_tK^{-1}P_t, S_{t-i}),
\end{equation}
where \texttt{interp} denotes the differentiable bilinear sampling operation, $K$ is the camera intrinsics matrix, and $S_{t-i \rightarrow t}$ is the semantic probility projected from $I_{t-i}$.

Thus, we can obtain the fused semantic information for keyframe $I_t$ from adjacent $k$ non-keyframes.
Since 2D segmentation is often accompanied with confidence, we fuse multi-view semantic information based on the normalized confidence weight with \texttt{Softmax} function: 
\begin{equation}
    S^m_{t} = \sum_{i}^{k} \texttt{softmax}(Conf_{t-i}) \cdot S_{t-i \rightarrow t},
\end{equation}
where $Conf_{t-i}$ is the confidence of $I_{t-i}$, and $S^m_{t}$ is the multi-view fused semantic probability, which can be used as the supervision of semantic MLP.

\subsection{Objective Functions}
\label{subsec:objective_function}
We perform camera tracking and bundle adjustment via objective function minimization. 
Like~\cite{azinovic:2022:neuralrgbd,vox-fusion}, to optimize the whole system, we apply five different objective functions: reconstruction loss, SDF loss, free space loss, semantic loss, and regularization loss.

\noindent\textbf{Reconstruction loss.}
For each ray with ground truth depth and color, we apply the reconstruction losses between the rendered value and ground truth value measured by the camera:
\begin{equation}
    \mathcal{L}_{c} =  \frac{1}{|R|} \sum_{r\in R}(\hat{c}(r) - I(r))^2, \quad \mathcal{L}_{d} = \frac{1}{|R|} \sum_{r\in R}(\hat{d}(r) - D(r))^2,
\end{equation}
where $R$ is the set of sampled pixels for optimization, \{$I(r)$,$D(r)$\} are the color and depth value of ray $r$, and \{$\hat{c}(r)$, $\hat{d}(r)$\} are the rendered value of sampled ray $r$, which are computed by~\cref{eq:render_propetry}.

\noindent\textbf{SDF and free space loss.}
Apart from the reconstruction loss, we also consider the geometrical information to supervise the reconstruction process.
To learn a better geometry representation, we apply constraints on the signed distance value, $\sigma$, with the following SDF loss and free space loss.
For the points near the surface and within the truncation region, \textit{e.g.} $|\hat{d}(p) - D(r)| < tr$, we use depth measurement to approximate the signed distance value:
\begin{equation}
    \mathcal{L}_{sdf} = \frac{1}{|R|} \sum_{r\in R} \frac{1}{|P_r^{tr}|}\sum_{p \in P_r^{tr}} (\hat{d}(p) + \sigma \cdot tr - D(r))^2,
\end{equation}
where $tr$ is the truncation distance, $P^{tr}_r$ is the set of sampled points within the truncation region.
The SDF loss $\mathcal{L}_{sdf}$ pushes the signed distance function to fit the surface of the scene.

Besides, for sampled points that are far away from the surface, \textit{e.g.} $|\hat{d}(p) - D(r)| > tr$, we apply a regularization on the free space that pushes its signed distance to 1:
\begin{equation}
    \mathcal{L}_{fs} = \frac{1}{|R|} \sum_{r\in R} \frac{1}{|P_r^{fs}|}\sum_{p \in P_r^{fs}} (\sigma - 1)^2,
\end{equation}
where $P_r^{fs}$ is the set of sampled points within the free space.
The free space loss $\mathcal{L}_{fs}$ pushes the signed distance value of empty space to 1, which reduces the noise or redundant points during the marching cubes.

\noindent\textbf{Semantic loss.}
To learn a consistent semantic field, we simultaneously use single-view predicted probability and the fused multi-view semantic probability as the supervision for the semantic decoder.
Here, we stop the gradient propagation from the semantic MLP to geometry MLP for stable geometry learning.
The semantic loss is defined as:
\begin{equation}
    \mathcal{L}_{sem} = - \frac{1}{|R|} \sum_{r\in R} \sum_{i=0}^{N_c} [(S_r(i) + \alpha \cdot S_r^m(i)) \log \hat{s}_r(i)],
\end{equation}
where $N_c$ is the number of classes for semantic segmentation, $S_r,S_r^m\in \mathbb{R}^{N_c}$ are the single-view and fused multi-view semantic probability for ray $r$, respectively. $\alpha$ is the weight parameter that balances two semantic components.

\noindent\textbf{Regularization loss.}
Unlike~\cite{sucar:2021:imap} the built-in smoothness priors, which uses a single MLP to represent the scene. 
To ensure the smoothness of the tetrahedron feature space and to avoid hash collisions when sampling vertex features, we apply regularization on the interpolated multi-resolution features:
\begin{equation}
    \mathcal{L}_{reg} = - \frac{1}{|R|} \sum_{r\in R} (\mathcal{F}(p;\Theta) - \mathcal{F}(p+\epsilon;\Theta))^2,
\end{equation}
where $\epsilon$ denotes the small perturbation of point location in 3D space, and $\mathcal{F}(\cdot ;\cdot)$ is the feature query functions in~\cref{subsec:representation}.
We only perform the regularization term of the multi-resolution tetrahedron feature during the mapping process on small random regions.

So, the final loss function is presented as follows:
\begin{equation}
\mathcal{L} = \lambda_{1} \mathcal{L}_{c} + \lambda_{2} \mathcal{L}_{depth} + \lambda_{3} \mathcal{L}_{fs} + \lambda_{4} \mathcal{L}_{sdf} + \lambda_{5} \mathcal{L}_{sem} + \lambda_{6} \mathcal{L}_{reg},
\end{equation}
where \{$\lambda_{i}$\} are the weight for each optimization component.

\subsection{Tracking}
\label{subsec:tracking}
For camera tracking, we fix the pose of the first frame as the identity matrix.
And for later coming frame $I_t$, we first initialize its camera-to-world pose, $T_{wc}^t=\exp(\xi)\in \mathbb{SE}(3)$, under the simple constant speed assumption.
Then we sample $M_t$ pixels across the frame for pose optimization.
During the tracking process, we fix the parameters of decoders and multi-resolution
tetrahedron features and only update $T_{wc}^t$.

\begin{figure}[ht!]
\centering
  \setlength{\tabcolsep}{1.5pt}
  \newcommand{\sz}{0.4}
  \begin{tabular}{cc}
    \makecell{\includegraphics[width=\sz\linewidth]{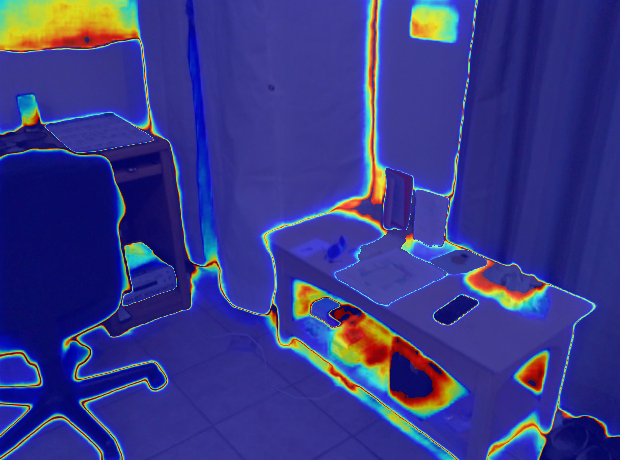}} &
    \makecell{\includegraphics[width=\sz\linewidth]{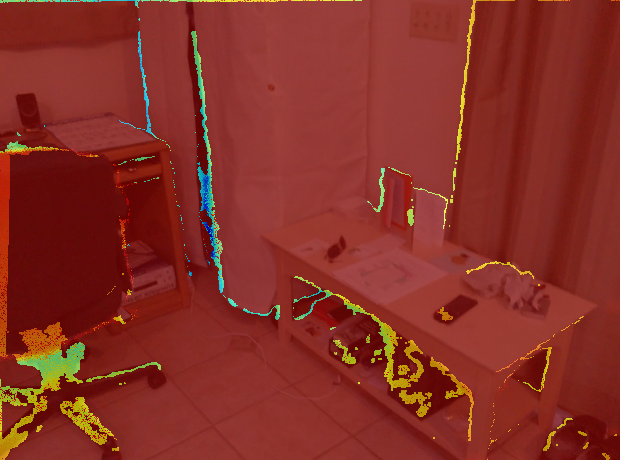}} \\
    \tt Confidence Map & \tt Tracking Error Map \\
  \end{tabular}
\caption{Segmentation confidence map and tracking error map. 
Complex geometry and ambiguous regions usually lead to lower confidence and slow convergence speed for color and geometry fields. As shown in the right figure, those regions will lead to large tracking errors.}
\vspace{-3mm}
\label{fig:conf_error}
\end{figure}

\noindent\textbf{Semantic-guided Pixel Sampling.}
For pixel sampling, we use confidence values from the semantic segmentation as the sampling probabilities for pixels.
Pixels with higher confidence values are deemed more reliable for camera tracking compared to those with lower confidence.
As shown in~\cref{fig:conf_error}, we show the confidence map (left part) and tracking error map of all pixels (right part).
Edge and challenge regions usually have low segmentation confidence and high tracking errors due to sensor noise and slow convergence speed of SDF filed in the complex geometry regions.
So, based on the confidence, we can avoid sampling these regions and use high-reliability and robust pixels to optimize the camera pose during tracking.

\noindent\textbf{Progressive Optimization Weight Function.}
As shown in~\cite{chen2023local2global,zhu2023latitude}, NeRF-based pose optimization is sensitive to the initialization.
The high-frequency information may lead to suboptimal results when the initial pose is far away from the ground truth.
Hence, we propose a progressive weight function that progressively updates the rendering weights during the iteration of the tracking process.  
Specifically, according to the render function in \cref{subsec:volume_rendering}, the distribution of rendering weight is controlled by the $\delta$.
Large $\delta$ means coarse constraints while small values represent fine-level constraints.
So, we apply a coarse-to-fine optimization loss for camera tracking and update $\delta$ in a decreasing manner via:
\begin{equation}
    \delta = \delta_{max} + (\delta_{min} - \delta_{max}) \sin(\frac{i}{N} \cdot \frac{\pi}{2}),
\end{equation}
where $i$ and $N$ are the current and total iteration steps for camera tracking, $\delta_{min}$, $\delta_{min}$ are the minimum and maximum values during optimization, and we use the $\sin(\cdot)$ function to update $\delta$ non-linearly.

Besides, for tracking, we set the weight of semantic loss as 0, because the predicted semantic results are noisy and inaccurate, and ground RGB and depth images are enough for camera tracking.

\subsection{Mapping}
\label{subsec:mapping}
We first initialize our permutohedral lattice features and MLP decoders with the first frame \{$I_0$,$D_0$\} for several iterations.
Then, we perform global bundle adjustments to update our model for every $k$ frame.
Following the practice in Co-SLAM~\cite{co-slam}, we use subset rays to construct the keyframe database instead of storing the whole keyframe~\cite{zhu:2021:niceslam}.
We randomly choose $M_m$ rays from the global keyframe database to perform bundle adjustments using all optimization functions in \cref{subsec:objective_function}.

\section{Experiments}
\label{sec:exp}
In this section, we introduce three commonly used datasets and our experimental setting.
When comparing with multiple baselines, we highlight the best three results as \colorbox{colorFst}{\bf first}, \colorbox{colorSnd}{second}, and \colorbox{colorTrd}{third} for better view.

\noindent\textbf{Dataset.}
We evaluate our approach on a variety of scenes from three different datasets.
For Replica dataset~\cite{julian:2019:replica}, we use the 8 synthetic scenes collected by iMAP~\cite{sucar:2021:imap} for camera tracking and reconstruction evaluation.
For the performance on real-world scenes, we use 5 scenes from ScanNet~\cite{dai:2017:scannet} and 3 scenes from TUM RGB-D dataset~\cite{sturm:2012:tumrgbd} for pose estimation.

\noindent\textbf{Implementation Details.}
Our geometry, color, and semantic decoders are implemented as MLP which consists of 2 fully connected (FC) layers with 32 latent dimensional channels.
The geometry decoder outputs a 15-dimensional latent feature vector and SDF scalar.
We use the ReLU activation function except for the final output.
Besides, we apply the Sigmoid function to the output of the color decoder, which limits the color value to $(0, 1)$. 
We use Adam optimizer and set hyper-parameters $beta = $(0.9, 0.999), $eps=1e-15$.
The learning rate for the parameters is set to 0.01.
We set the truncation distance $tr$ to 5cm in our method.
For the Replica dataset, we use 40 and 100 iterations for tracking and mapping.
For the ScanNet dataset, we use 40 and 50 iterations for tracking and mapping.
The loss weights for different optimization components are $\lambda_1$=5.0, $\lambda_2$=0.1, $\lambda_3$=1e3, $\lambda_4$=10, $\lambda_5$=1.0, $\lambda_6$=1e-6.
We use~\cite{rosu2023permutosdf} to encode the tetrahedron feature, and the multi-resolution tetrahedron features contain 16 level details, each containing a feature vector of 2 dimensions.
For positional encoding, we set the frequency to 8, which leads to a 48-dimensional feature vector.
So, the final input feature of our system is the concatenated 80 channels feature.
The finest tetrahedron resolution is set to 2cm.

\begin{table}[tb]
\centering
\setlength{\tabcolsep}{2pt}
\renewcommand{\arraystretch}{1.05}
\resizebox{\columnwidth}{!}
{
\begin{tabular}{lccccccccc}
\toprule
Method & \texttt{Rm0} & \texttt{Rm1} & \texttt{Rm2} & \texttt{Off0} & \texttt{Off1} & \texttt{Off2} & \texttt{Off3} & \texttt{Off4} & Avg.\\
\midrule
\multirow{1}{*}{NICE-SLAM~\cite{zhu:2021:niceslam}} &  0.97
& 1.31 & 1.07  &  0.88 &  1.00 & 1.06  & 1.10  & 1.13 & 1.06 \\[0.8pt] \noalign{\vskip 1pt}
\multirow{1}{*}{Vox-Fusion~\cite{vox-fusion}} &  \nd 0.40 &  \rd 0.54 &  0.54 & \rd 0.50 & \nd 0.46 &  \rd 0.75 & \fs 0.50 &  \rd 0.60 & \rd 0.54 \\ [0.8pt] \noalign{\vskip 1pt}
%
\multirow{1}{*}{Co-SLAM~\cite{co-slam}} & 0.77 & 1.04 & 1.09 & 0.58 & 0.53 & 2.05 & 1.49 & 0.84 & 0.99 \\[0.8pt]  \noalign{\vskip 1pt}
\multirow{1}{*}{ESLAM~\cite{eslam}} &  0.71 & 0.70 & \rd 0.52  &  0.57  &  0.55  & \nd 0.58  & 0.72 & 0.63 & 0.63 \\[0.8pt] \noalign{\vskip 1pt}%
\multirow{1}{*}{Point-SLAM~\cite{point-slam}} &  \rd 0.61  & \nd 0.41  & \nd 0.37  & \nd 0.38 & \rd 0.48 & \fs 0.54  & \rd 0.69 & \rd 0.72 & \nd 0.52 \\[0.8pt]  \noalign{\vskip 1pt}
Ours & \fs 0.30	& \fs 0.40 & \fs 0.36 & \fs 0.29 & \fs 0.31 & 0.92 & \nd 0.67 & \fs 0.44 & \fs 0.46 \\
\bottomrule
\end{tabular}
}
\caption{Tracking Performance on Replica~\cite{julian:2019:replica} (ATE RMSE $\downarrow$ [cm]). The numbers of compared methods are taken from the original papers or~\cite{point-slam,co-slam}.}
\label{tab:pose_replica}
\end{table}

\begin{table}[tp]
\centering
\resizebox{1.00\columnwidth}{!}{
\begin{tabular}{lcccccc}
\toprule
Method & \texttt{0000} & \texttt{0059} & \texttt{0106} &  \texttt{0169}  & \texttt{0207} & Avg. \\
\midrule 
DI-Fusion~\cite{huang2021di-fusion} & 62.99 & 128.00 & 18.50 & 75.80 & 100.19 & 77.10 \\ [0.8pt] \noalign{\vskip 1pt}
iMAP~\cite{sucar:2021:imap} & 55.95 & 32.06 & 17.50 & 70.51  & 11.91 & 37.59 \\[0.8pt] \noalign{\vskip 1pt}
NICE-SLAM~\cite{zhu:2021:niceslam}  & 12.00 &  14.00 & \nd 7.90 & 10.90 & \nd 6.20 &  10.20 \\[0.8pt] \noalign{\vskip 1pt}
Vox-Fusion~\cite{vox-fusion} & \nd 8.39 & \nd 9.13 & \fs 7.44 & \nd 6.53 & \fs 5.57 & \fs 7.41 \\[0.8pt] \noalign{\vskip 1pt}
Co-SLAM~\cite{co-slam} & \fs 7.18 & 12.29 & 9.57  & \rd 6.62 & 7.13 & \rd 8.56 \\ [0.8pt] \noalign{\vskip 1pt}
Point-SLAM~\cite{point-slam} & 10.24  & \fs 7.81  &  8.65  & 22.16 & 9.54 & 11.68 \\ [0.8pt] \noalign{\vskip 1pt}
Ours &  \rd 8.70  & \rd 9.62 & \rd 8.35 &  \fs 5.64  & \rd 7.10 & \nd 7.88 \\ [0.8pt] \noalign{\vskip 1pt}
        
\bottomrule
\end{tabular}
}
\caption{Tracking Performance on ScanNet~\cite{dai:2017:scannet} (ATE RMSE $\downarrow$ [cm]). The numbers of all compared methods are taken from the original papers or\cite{point-slam,co-slam}. The value of Vox-Fusion~\cite{vox-fusion} on \texttt{0059} is obtained with their released code. All scenes are evaluated on the \texttt{00} trajectory.}
\label{tab:pose_scannet}
\end{table}

\begin{table}[tb]
\centering
 \resizebox{1.00\columnwidth}{!}{
\begin{tabular}{lcccc}
\toprule
Method & \texttt{fr1/desk} & \texttt{fr2/xyz} & \texttt{fr3/office}  & Avg.\\
\midrule
DI-Fusion~\cite{huang2021di-fusion} & 4.4   & 2.0 & 5.8 &  4.06\\
NICE-SLAM~\cite{zhu:2021:niceslam} & 4.26  & 6.19  &3.87 & 4.77 \\
Vox-Fusion~\cite{vox-fusion}  & \rd 3.52 & \nd 1.49 &  26.01 &  10.34 \\
Co-SLAM~\cite{co-slam} & \nd 2.70  & 1.90  & \nd 2.60 &  \rd 2.50 \\
Point-SLAM~\cite{point-slam} & 4.34  & \fs 1.31  & \rd 3.48 & \nd 2.40 \\
Ours & \fs 2.27  & \rd 1.63  & \fs 2.36 &  \fs 2.08 \\
\bottomrule
\end{tabular}
}
\caption{Tracking Performance on TUM RGB-D~\cite{sturm:2012:tumrgbd} (ATE RMSE $\downarrow$ [cm]). The numbers of compared methods are taken from \cite{point-slam,co-slam}.}
\label{tab:pose_tum}
\end{table}

\begin{figure*}[t]
  \centering
  \footnotesize
  \setlength{\tabcolsep}{1.5pt}
  \newcommand{\sz}{0.235}
  \begin{tabular}{lcccc}
    & NICE-SLAM~\cite{zhu:2021:niceslam} & Point-SLAM~\cite{point-slam} & Ours & GT \\
    \makecell{\rotatebox{90}{\tt scene0000}} &
    \makecell{\includegraphics[width=\sz\linewidth]{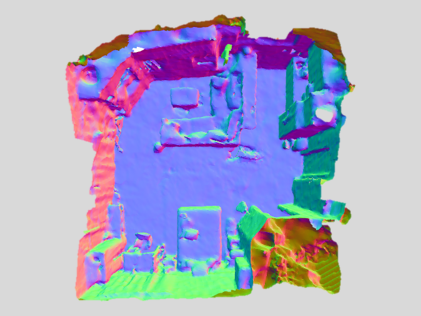}} &
    \makecell{\includegraphics[width=\sz\linewidth]{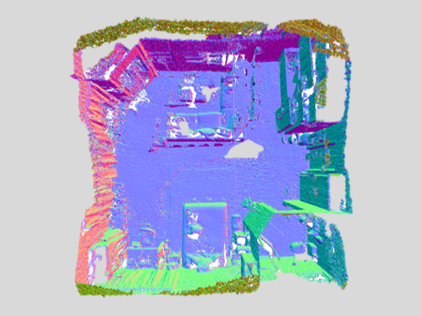}} &
    \makecell{\includegraphics[width=\sz\linewidth]{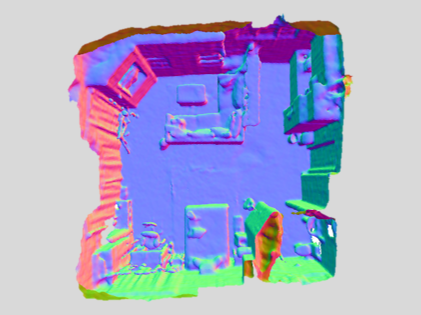}} &
    \makecell{\includegraphics[width=\sz\linewidth]{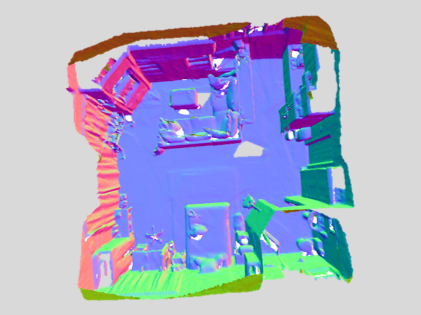}} \\
    \makecell{\rotatebox{90}{\tt scene0106}} &
    \makecell{\includegraphics[width=\sz\linewidth]{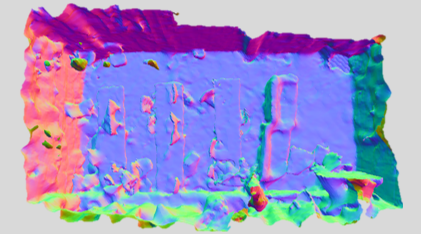}} &
    \makecell{\includegraphics[width=\sz\linewidth]{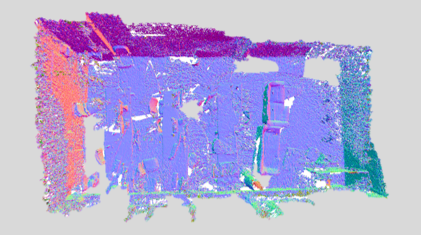}} &
    \makecell{\includegraphics[width=\sz\linewidth]{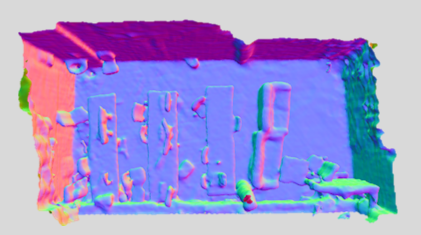}} &
    \makecell{\includegraphics[width=\sz\linewidth]{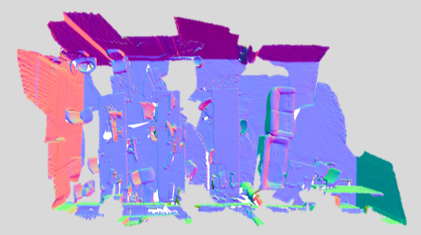}} \\
    \makecell{\rotatebox{90}{\tt scene0169}} &
    \makecell{\includegraphics[width=\sz\linewidth]{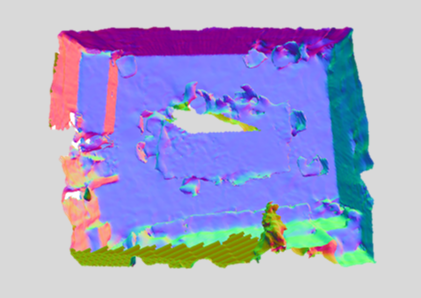}} &
    \makecell{\includegraphics[width=\sz\linewidth]{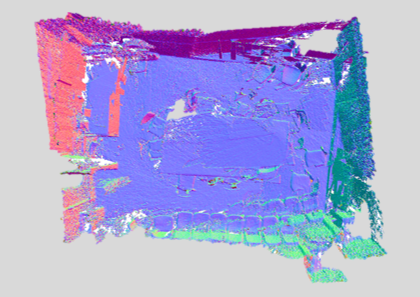}} &
    \makecell{\includegraphics[width=\sz\linewidth]{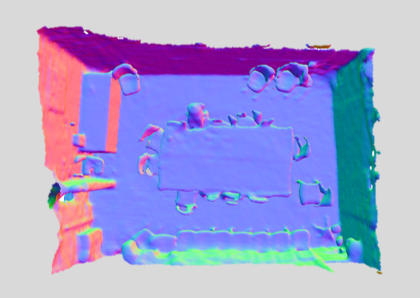}} &
    \makecell{\includegraphics[width=\sz\linewidth]{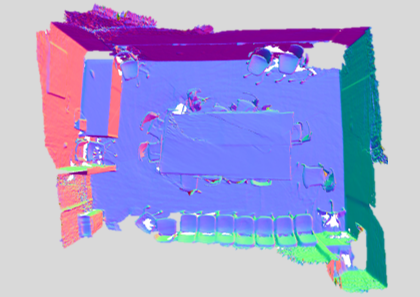}} \\
  \end{tabular} 
 
  \caption{Reconstruction Results on ScanNet~\cite{dai:2017:scannet}. Compared to the baselines, our method can reconstruct more accurate detailed geometry and generate more complete, smoother mesh.}
  \label{fig:scannet_mesh}
\end{figure*}
\begin{figure*}[t]
  \centering
  \scriptsize
  \setlength{\tabcolsep}{0.5pt}
  \newcommand{\sz}{1.8}  %
  \begin{tabular}{lc}
    \makecell{\rotatebox{90}{vMAP~\cite{kong2023vmap}}}  &
    \makecell{\includegraphics[width=\sz\columnwidth]{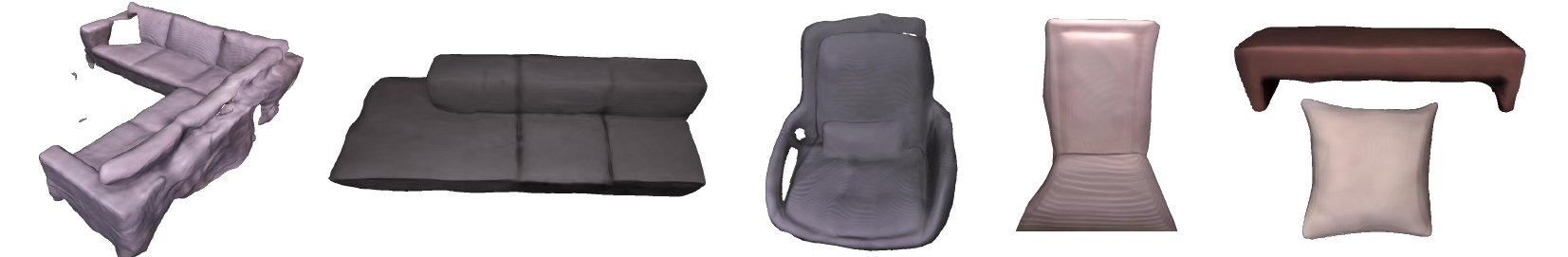}} \\
    \makecell{\rotatebox{90}{Ours}}  &
    \makecell{\includegraphics[width=\sz\columnwidth]{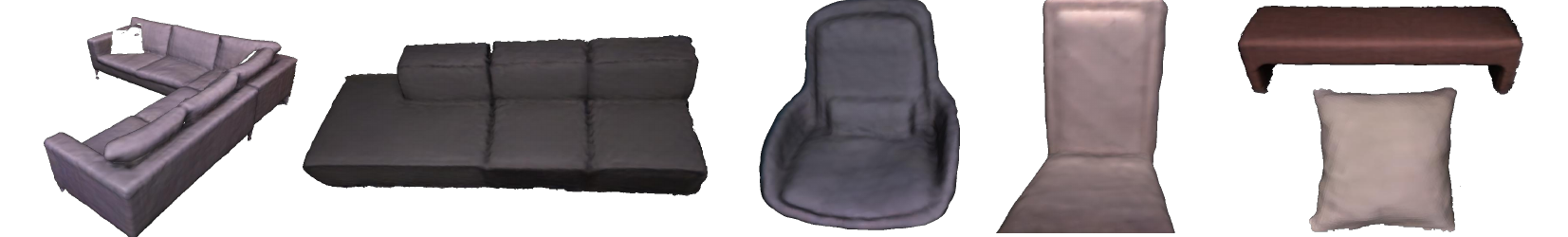}} \\
    \end{tabular}
  \caption{Object reconstruction of Replica~\cite{julian:2019:replica}. We show some selected objects for comparison with vMAP~\cite{kong2023vmap}.}
  \label{fig:object_meshs}
\end{figure*}

\subsection{Evaluation of Camera Tracking}
In this part, we evaluate the performance of camera tracking on Replica~\cite{julian:2019:replica}, ScanNet~\cite{dai:2017:scannet}, and TUM RGB-D~\cite{sturm:2012:tumrgbd} datasets.
The compared baseline methods, evaluation metrics, and results of this part are shown in the following.

\noindent\textbf{Baselines and Metrics.}
Following the setting in~\cite{point-slam,co-slam}, we take iMAP~\cite{sucar:2021:imap}, NICE-SLAM~\cite{zhu:2021:niceslam}, Vox-Fusion~\cite{vox-fusion}, Co-SLAM~\cite{co-slam}, ESLAM~\cite{eslam}, and Point-SLAM~\cite{point-slam} as the compared baselines.
In order to measure the camera tracking accuracy of the proposed method,
we use the commonly absolute trajectory error ATE RMSE~\cite{sturm2012benchmark} as the evaluation metric.
We align the estimated trajectory and ground truth trajectory using Horn`s closed form solution~\cite{horn1988closed} for a fair comparison.

\noindent\textbf{Qualitative and Quantitative Results.}
We show the camera tracking performance on the Replica dataset in \cref{tab:pose_replica}.
As shown by the results, we achieve the best tracking performance across most scenarios, except for the \texttt{Office 2} scene. Overall, our method outperforms the compared baselines in terms of average results.
Besides, to validate the performance on real-world scenes, we conduct experiments on ScanNet~\cite{dai:2017:scannet} and TUM-RGBD~\cite{sturm:2012:tumrgbd} datasets.
The results are shown in \cref{tab:pose_scannet} and \cref{tab:pose_tum}, respectively.
For TUM-RGBD dataset, only the sequence \texttt{fr3/xyz} achieved third performance. On the rest sequences, our approach achieves the best results, including average data.
For ScanNet dataset, our camera tracking performance is slightly worse than Vox-Fusion. The reason why Vox-Fusion can achieve good results is that it performs dense BA for each frame. The running speed of Vox-Fusion is rather slow ($\sim$3s per frame).
Real-world scenarios often involve complex environments, including depth noise, motion blur, and non-Lambertian materials, which makes the neural radiance field hard to model. 
Compared to the point or voxel based approaches, our performance of reconstructing high-frequency color information will be slightly worse due to the lack of explicit representation.


\begin{table*}[t]
\centering
\footnotesize
\begin{tabularx}{\linewidth}{llccccccccc}
\toprule
Method & Metric & \texttt{Room0} & \texttt{Room1} & \texttt{Room2} & \texttt{Office0} & \texttt{Office1} & \texttt{Office2} & \texttt{Office3} & \texttt{Office4} & Avg.\\
\midrule
\multirow{3}{*}{iMAP~\cite{sucar:2021:imap}} & \textbf{Acc.}[cm]$\downarrow$ & 3.58 & 3.69 & 4.68 & 5.87 & 3.71 & 4.81 & 4.27 & 4.83 & 4.43 \\
                      & \textbf{Comp.}[cm]$\downarrow$ & 5.06 & 4.87 & 5.51 & 6.11 & 5.26 & 5.65 & 5.45 & 6.59 & 5.56 \\
                      & \textbf{Comp. Ratio}[$<5$cm \%]$\uparrow$ & 83.91 & 83.45 & 75.53 & 77.71 & 79.64 & 77.22 & 77.34 & 77.63 & 79.06\\
\hdashline
\multirow{3}{*}{\makecell[l]{NICE-SLAM~\cite{zhu:2021:niceslam}}} & \textbf{Acc.}[cm]$\downarrow$ & 2.97 & 3.23 & 3.46 & 5.47 & 3.33 & 4.40 & 3.55 & 2.87 & 3.66 \\
                      & \textbf{Comp.}[cm]$\downarrow$ & 3.30 &  3.07 & 3.75 & 4.54 & 3.83 & 3.90 & 4.49 & 3.91 & 3.85 \\
                      & \textbf{Comp. Ratio}[$<5$cm \%]$\uparrow$ & 89.51 & 86.01 & 81.14 & 85.27 & 88.01 & 82.61 & 79.49 & 85.33 & 84.67 \\

\hdashline
\multirow{3}{*}{\makecell[l]{Co-SLAM~\cite{co-slam}}} & \textbf{Acc.}[cm]$\downarrow$ & \rd 1.61 & \rd 1.31 & \rd 1.55 & \rd 1.33 & \rd 1.11 & \rd 1.83 & \rd 1.97 & \rd 1.73 & \rd 1.56 \\
                      & \textbf{Comp.}[cm]$\downarrow$ & \nd 2.96 & \nd 2.46 & \nd 2.36 & \nd 1.43 & \nd 1.82 & \nd 3.26 & \nd 3.26 & \nd 3.36 & \nd 2.61 \\
                      & \textbf{Comp. Ratio}[$<5$cm \%]$\uparrow$ & \nd 91.12 & \nd 92.18 & \nd 91.44 & \nd 95.65 & \nd 93.56 & \nd 88.53 & \nd 87.67 & \nd 87.76 & \nd 90.99 \\
\hdashline
\multirow{3}{*}{\makecell[l]{Point-SLAM~\cite{point-slam}}} & \textbf{Acc.}[cm]$\downarrow$ & \fs 1.45 & \fs 1.14 & \fs 1.19 & \fs 1.05 & \fs 0.86 & \fs 1.31 & \fs 1.57 & \fs 1.51 & \fs 1.26 \\
                      & \textbf{Comp.}[cm]$\downarrow$ & \rd 3.46 & \rd  3.02 & \rd 2.65 & \rd 1.65 & \rd 2.21 & \rd 3.62 & \rd 3.47 & \rd 3.90 & \rd 3.00 \\
                      & \textbf{Comp. Ratio}[$<5$cm \%]$\uparrow$ & \rd 88.48 & \rd 89.44 & \rd 90.13 & \rd 93.39 & \rd 90.51 & \rd 86.17 & \rd 86.00 & \rd 85.74 & \rd 88.73 \\
\hdashline
\multirow{3}{*}{Ours} & \textbf{Acc.}[cm]$\downarrow$ & \nd 1.54 & \nd 1.25 & \nd 1.41 & \nd 1.20 & \nd 1.09 & \nd 1.58 & \nd 2.16 & \nd 1.60 & \nd 1.48 \\
                      & \textbf{Comp.}[cm]$\downarrow$ & \fs 2.81 & \fs 2.37 & \fs 2.23 & \fs 1.37 & \fs 1.67 & \fs 2.86 & \fs 2.98 & \fs 3.24 & \fs 2.44 \\
                      & \textbf{Comp. Ratio}[$<5$cm \%]$\uparrow$ & \fs 92.14 & \fs 92.88	& \fs 92.98 & \fs 97.13 & \fs 95.08 & \fs 90.87 & \fs 90.17 & \fs 88.67 &	\fs 92.49 \\
\bottomrule
\end{tabularx}
\caption{Reconstruction Performance on Replica~\cite{julian:2019:replica}. For NICE-SLAM and iMAP, we take the number from~\cite{zhu:2021:niceslam}. For Point-SLAM~\cite{point-slam} and Co-SLAM~\cite{co-slam}, we generate the mesh with their open-source codes.}
\label{tab:recon_replica}
\end{table*}

\begin{table*}[!tp]
\centering
\footnotesize
\begin{tabularx}{\linewidth}{llccccccccc}
\toprule
Method & Metric & \texttt{Room0} & \texttt{Room1} & \texttt{Room2} & \texttt{Office0} & \texttt{Office1} & \texttt{Office2} & \texttt{Office3} & \texttt{Office4} & Avg.\\
\midrule
\multirow{3}{*}{vMAP~\cite{kong2023vmap}} & \textbf{Object Acc.}[cm]$\downarrow$ & \nd 3.50 & \nd 3.06 & \nd 3.37 & \nd 2.45 & \nd 3.04 & \nd 3.07 & \nd 3.18 & \nd 3.07 & \nd 3.09 \\
                      & \textbf{Object Comp.}[cm]$\downarrow$ & \nd 3.10 & \nd 3.84 & \nd 7.11 & \nd 3.43 & \nd 1.47 & \nd 3.93 & \nd 4.94 & \nd 9.33 & \nd 4.64 \\
                      & \textbf{Object Comp. Ratio}[$<5$cm \%]$\uparrow$ & \nd 84.02 & \nd 88.32 & \nd 61.84 & \nd 69.66 & \nd 94.58 & \nd 82.52 & \nd 77.97 & \nd 62.45 & \nd 77.67\\
\hdashline
\multirow{3}{*}{Ours} & \textbf{Object Acc.}[cm]$\downarrow$ & \fs 2.05 & \fs 1.47 & \fs 2.87 & \fs 1.96 & \fs 1.27 & \fs 1.84 & \fs 2.15 & \fs 2.76 & \fs 2.04 \\
                      & \textbf{Object Comp.}[cm]$\downarrow$ & \fs 3.50 & \fs 3.37 & \fs 4.87 & \fs 2.62 & \fs 1.39 & \fs 3.46 & \fs 4.41 & \fs 7.87 & \fs 3.94 \\
                      & \textbf{Object Comp. Ratio}[$<5$cm \%]$\uparrow$ & \fs 80.75 & \fs 87.75 & \fs 73.21 & \fs 88.05 & \fs	95.29 & \fs 85.27 & \fs 76.12 & \fs	65.92 & \fs	81.52 \\
\bottomrule
\end{tabularx}
\caption{Object Reconstruction Performance on Replica~\cite{julian:2019:replica}. 
We reconstruct the object with the pose estimate by ourselves.
The numbers of vMAP~\cite{kong2023vmap} are evaluated on the meshes obtained from their code with the same pose of ours.}
\label{tab:replica_object_recon}
\end{table*}

\begin{figure*}[ht]
  \centering
  \scriptsize
  \setlength{\tabcolsep}{0.5pt}
  \newcommand{\sz}{0.22}  %
  \begin{tabular}{lccccc}
    \makecell{\rotatebox{90}{Mask2Former~\cite{cheng2021mask2former}}}  &
    \makecell{\includegraphics[height=\sz\columnwidth]{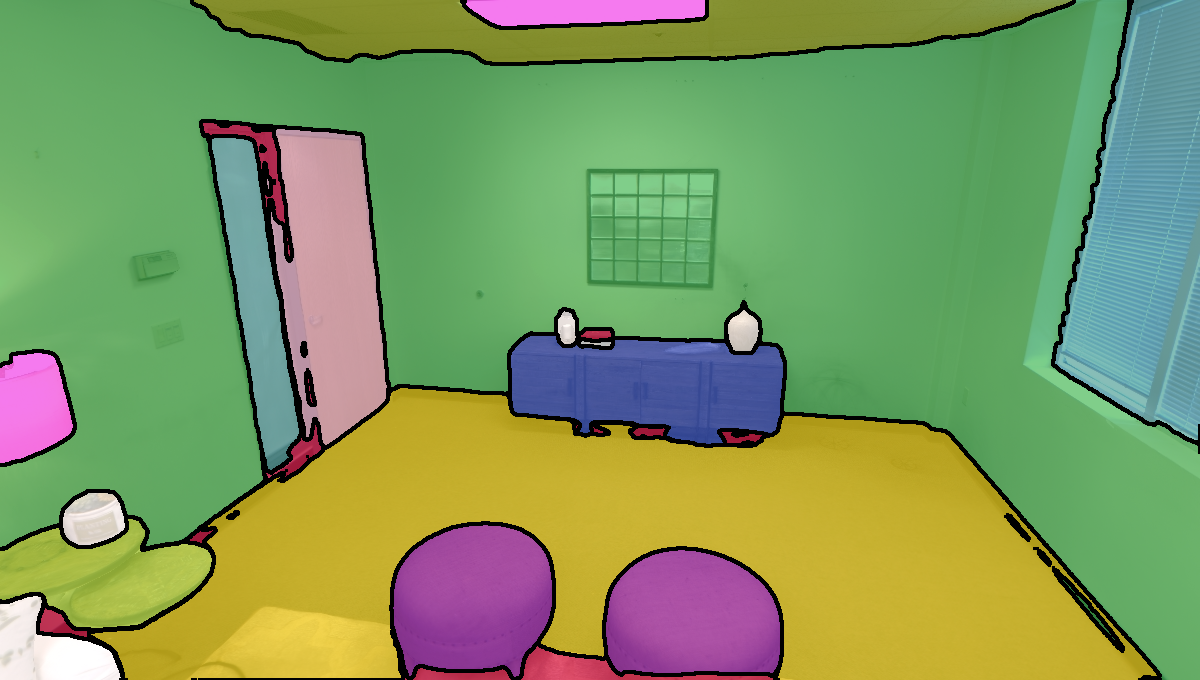}} & 
    \makecell{\includegraphics[height=\sz\columnwidth]{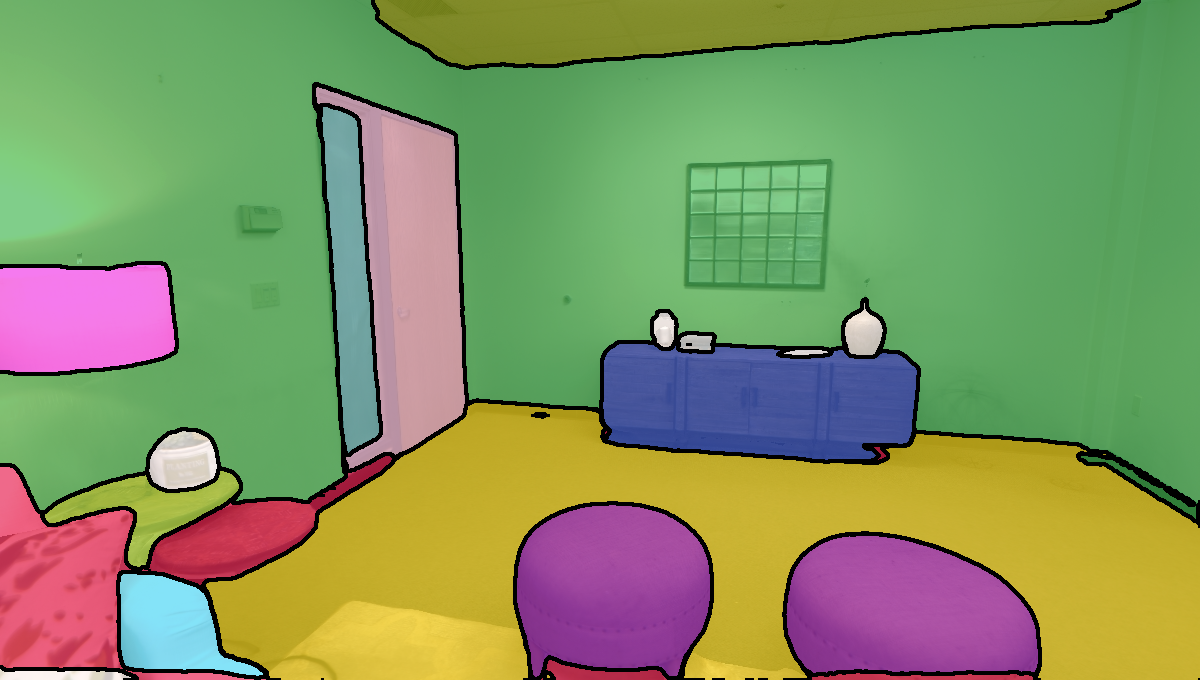}} & 
    \makecell{\includegraphics[height=\sz\columnwidth]{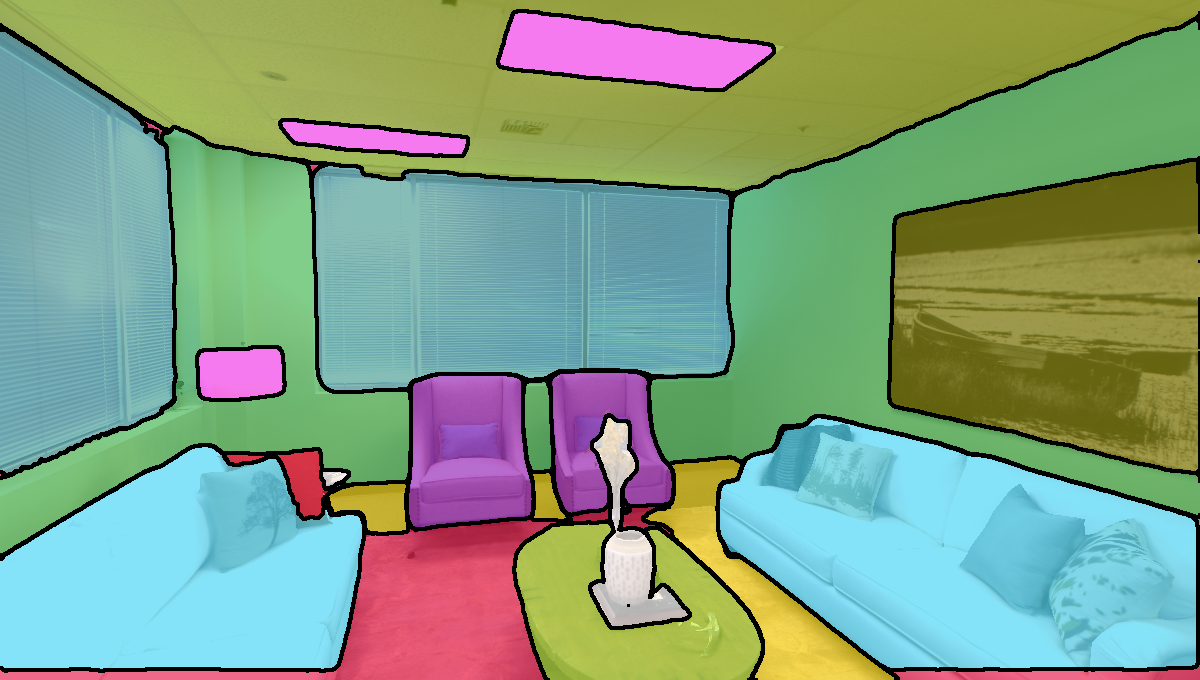}} & 
    \makecell{\includegraphics[height=\sz\columnwidth]{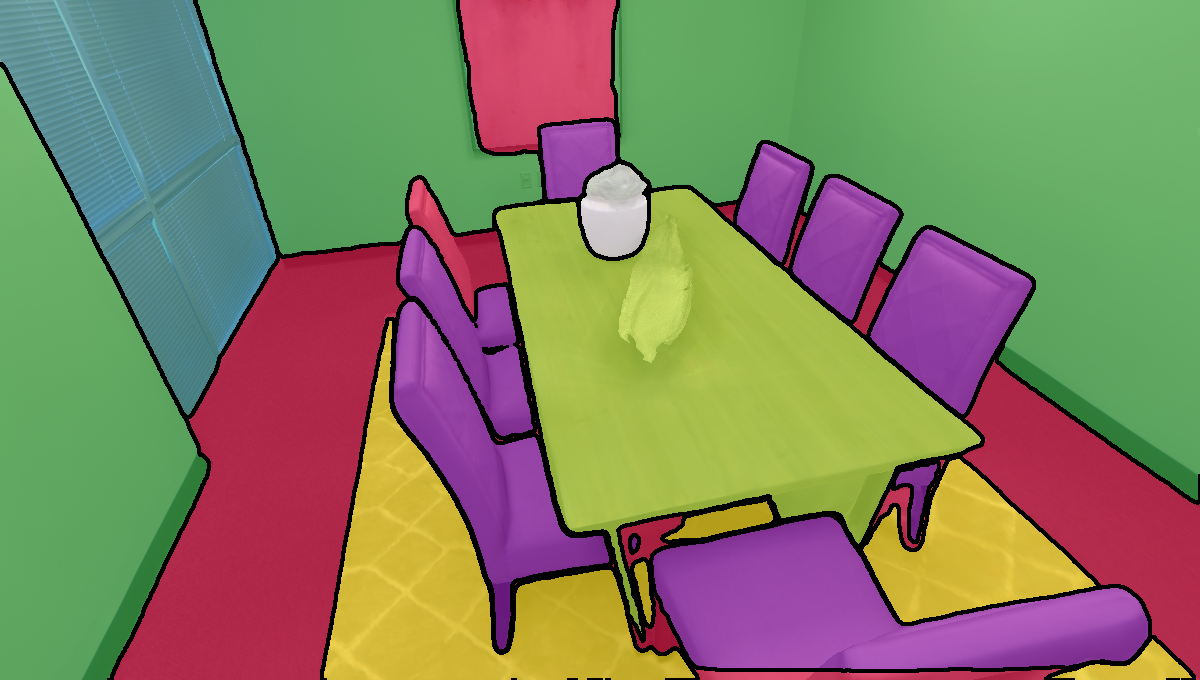}} & 
    \makecell{\includegraphics[height=\sz\columnwidth]{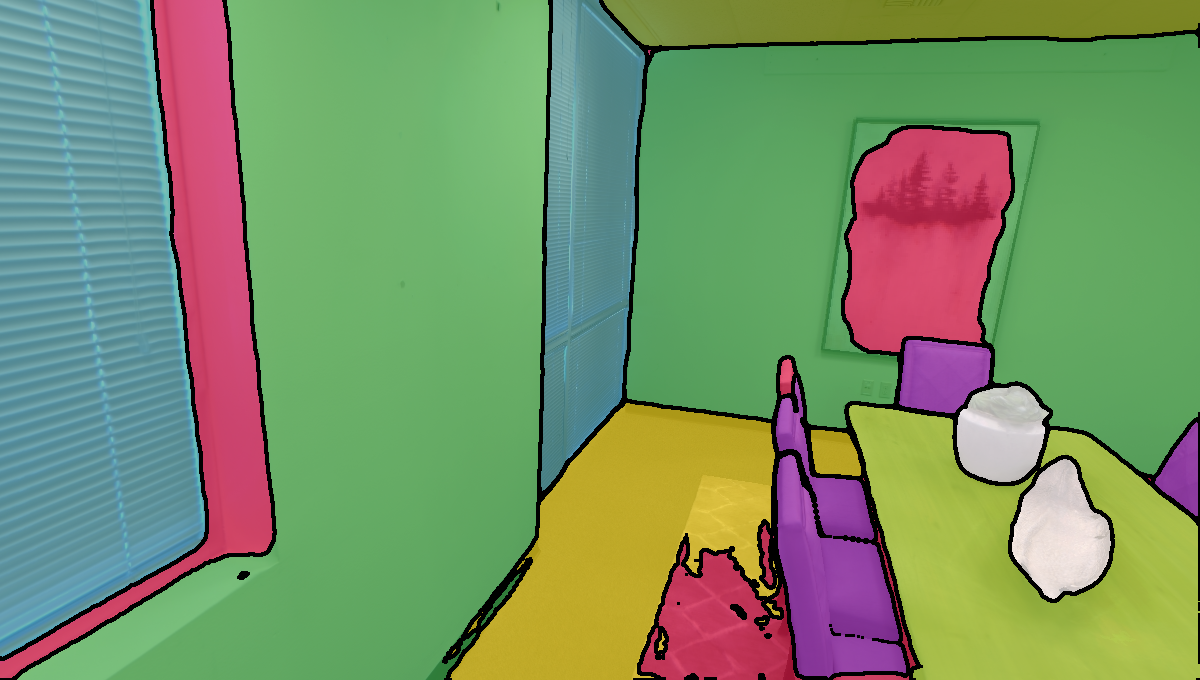}} \\ 
    \makecell{\rotatebox{90}{No Fusion}}  &
    \makecell{\includegraphics[height=\sz\columnwidth]{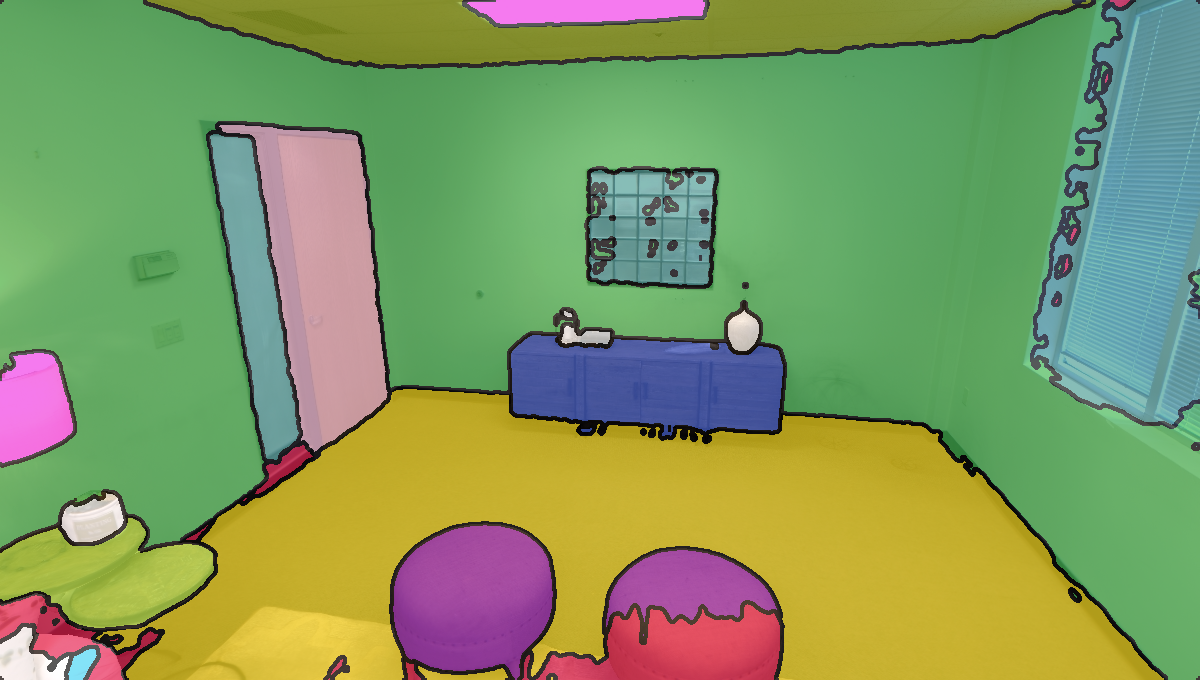}} & 
    \makecell{\includegraphics[height=\sz\columnwidth]{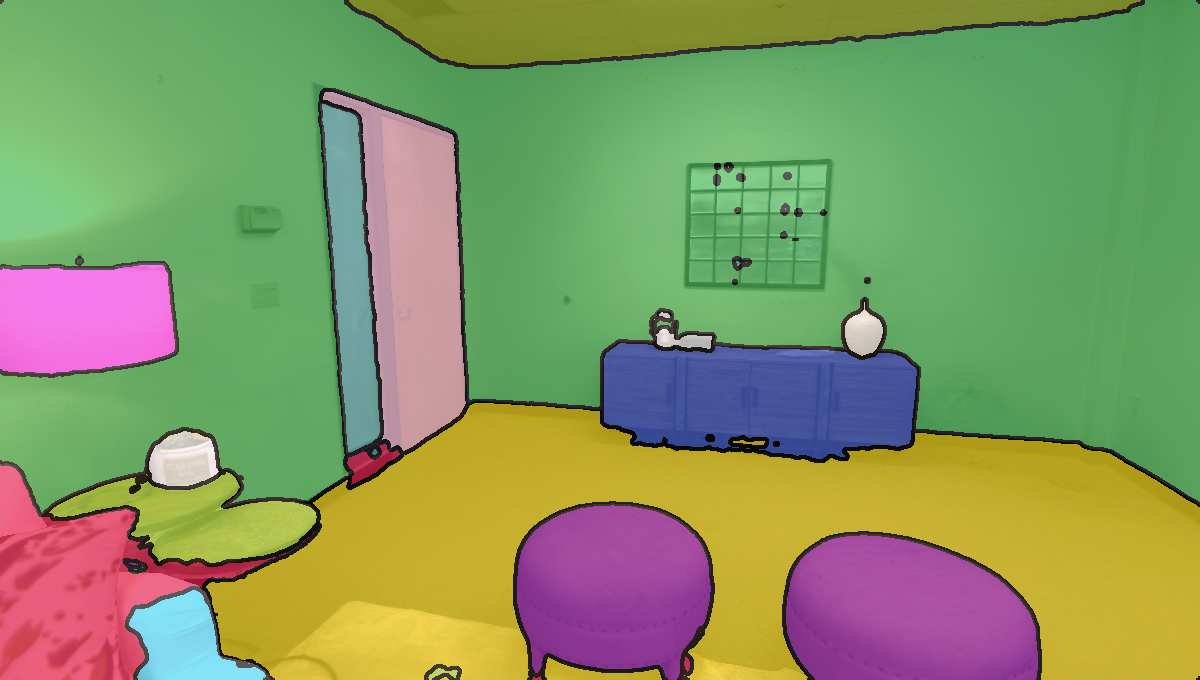}} & 
    \makecell{\includegraphics[height=\sz\columnwidth]{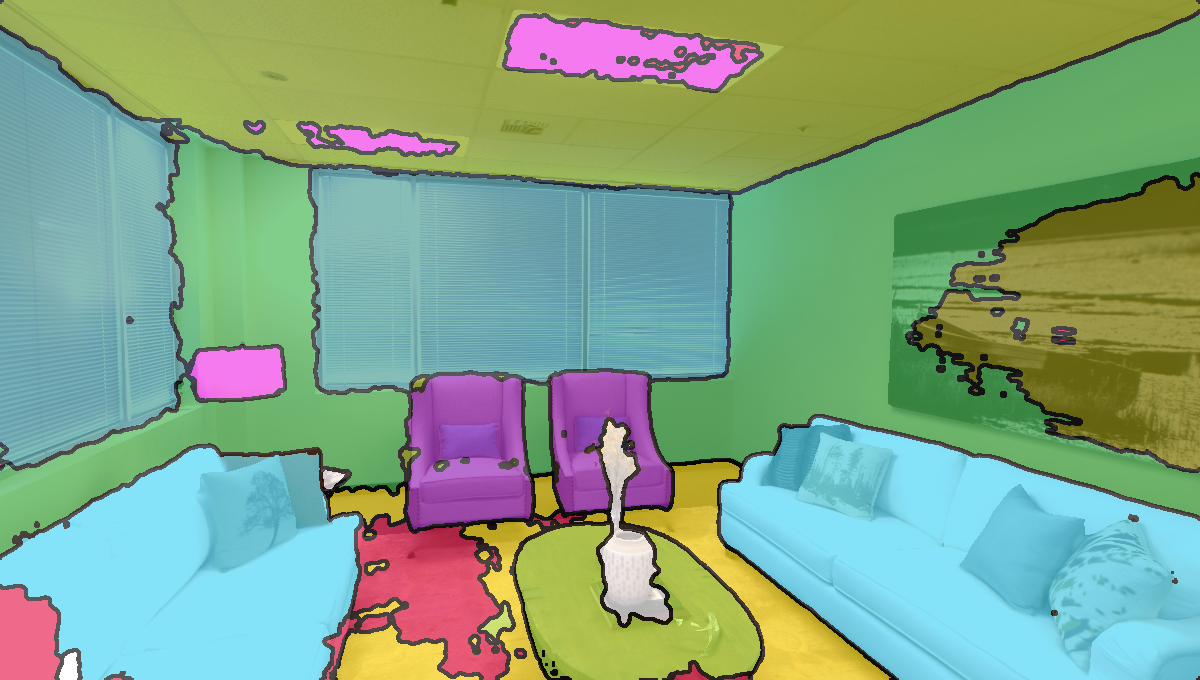}} & 
    \makecell{\includegraphics[height=\sz\columnwidth]{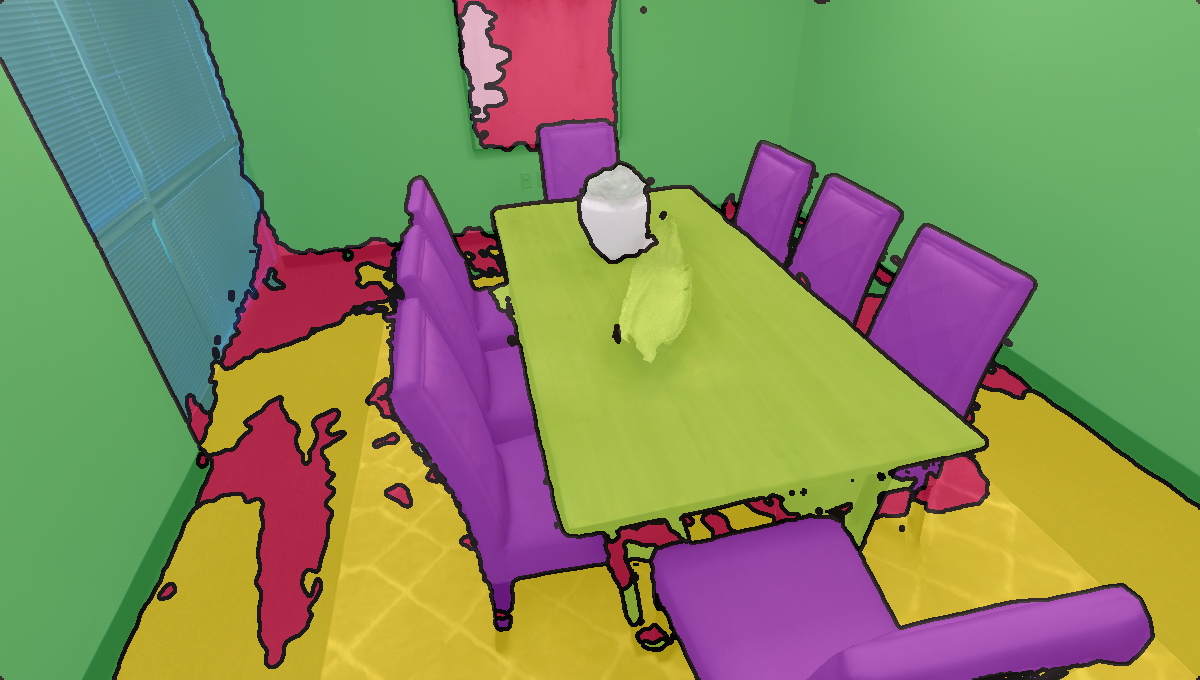}} & 
    \makecell{\includegraphics[height=\sz\columnwidth]{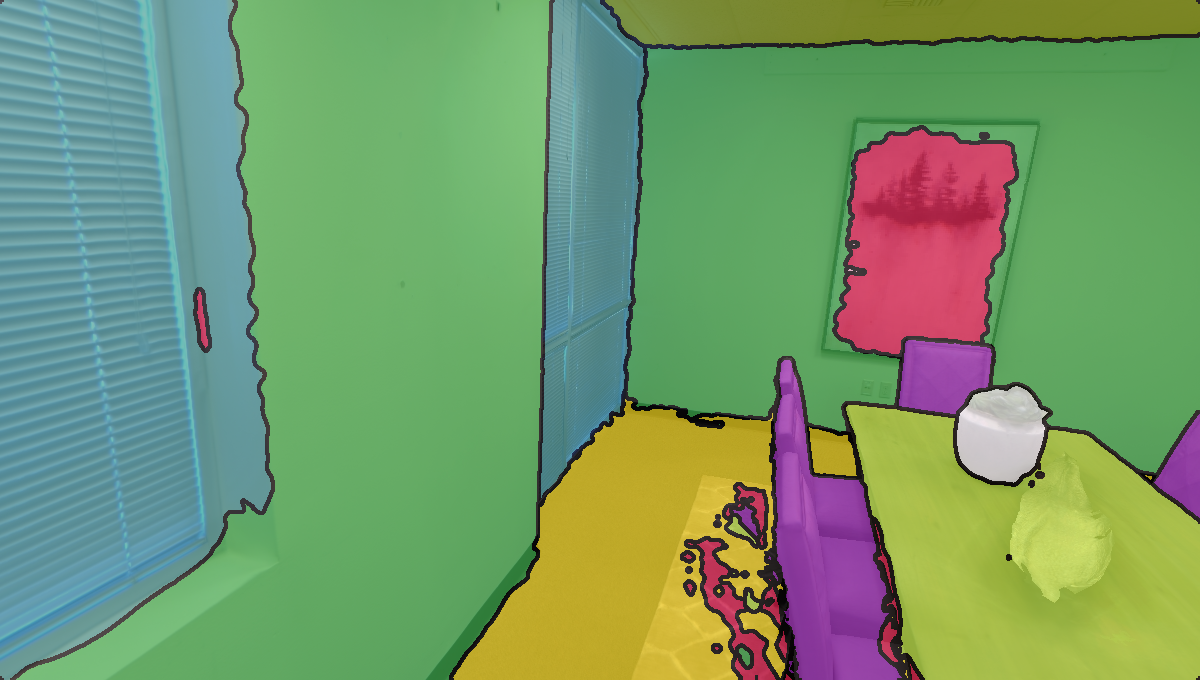}} \\ 
    \makecell{\rotatebox{90}{Ours}}  &
    \makecell{\includegraphics[height=\sz\columnwidth]{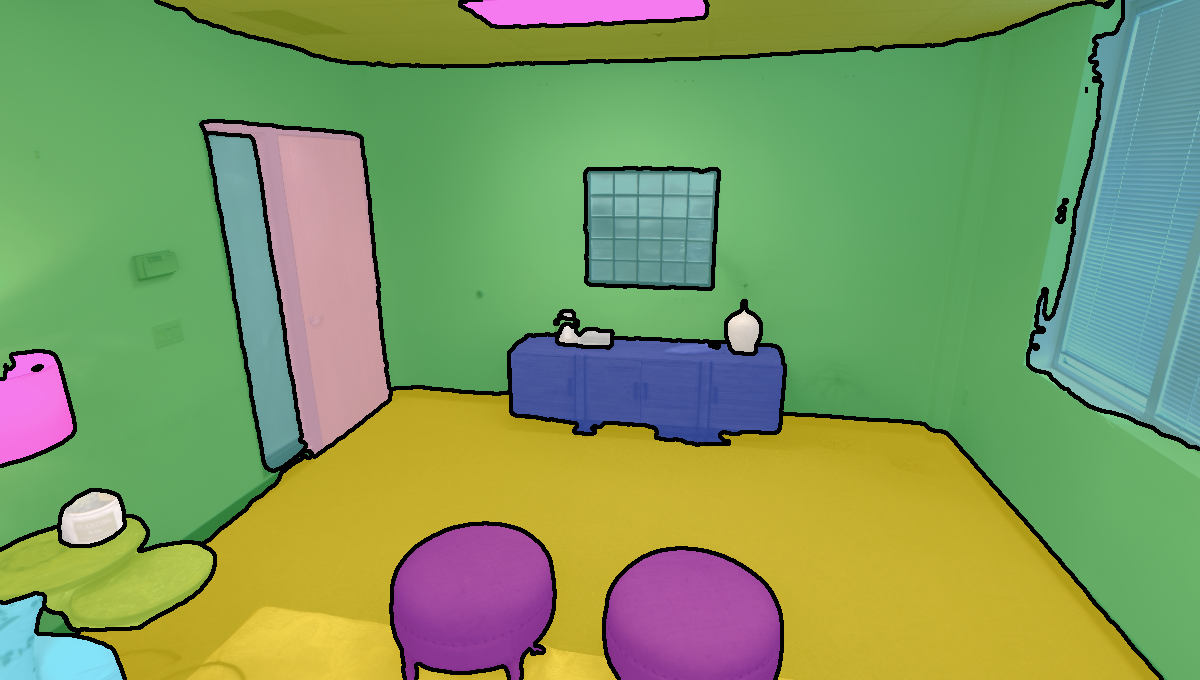}} & 
    \makecell{\includegraphics[height=\sz\columnwidth]{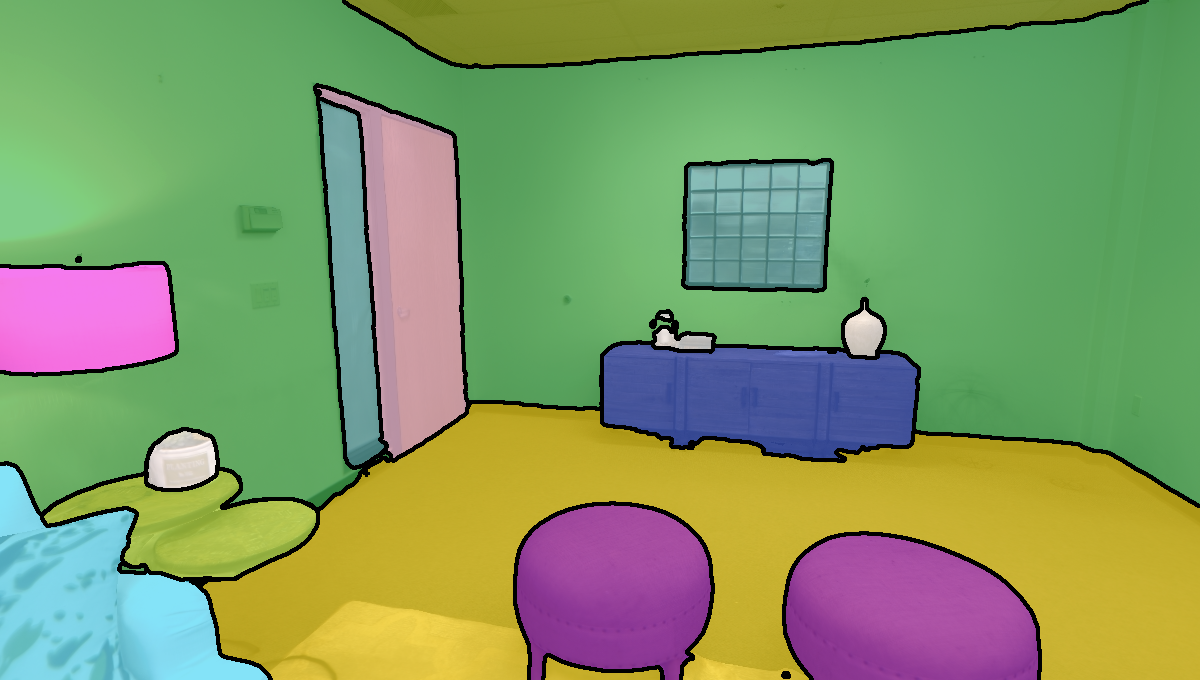}} & 
    \makecell{\includegraphics[height=\sz\columnwidth]{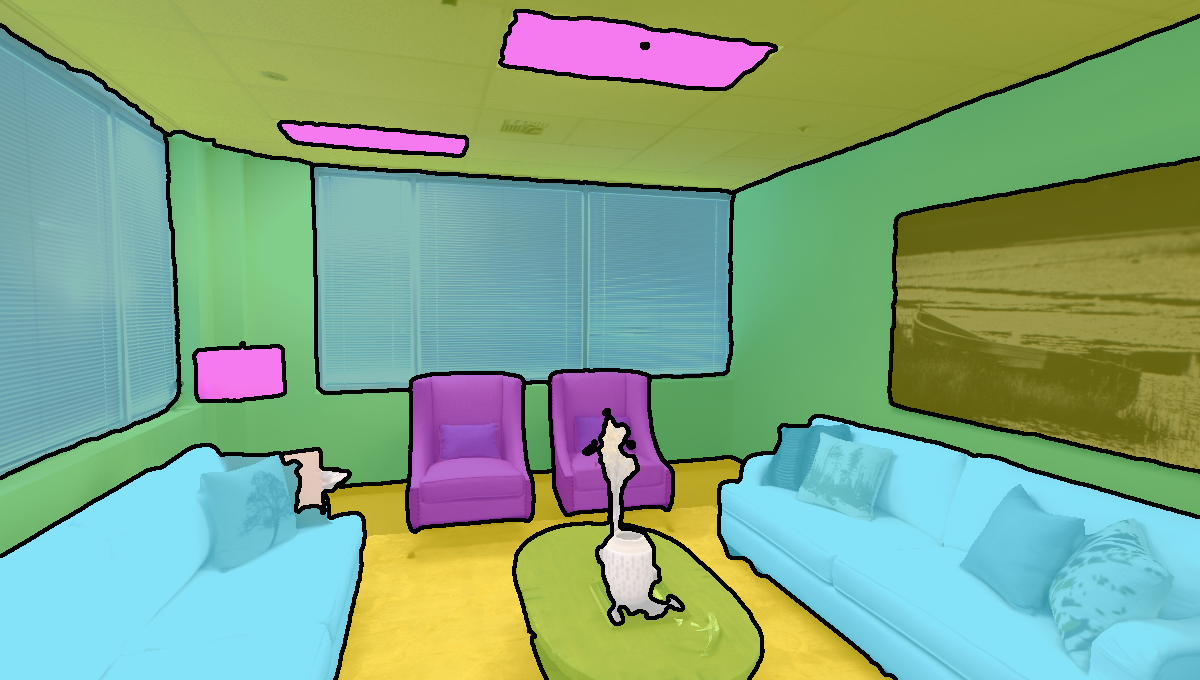}} & 
    \makecell{\includegraphics[height=\sz\columnwidth]{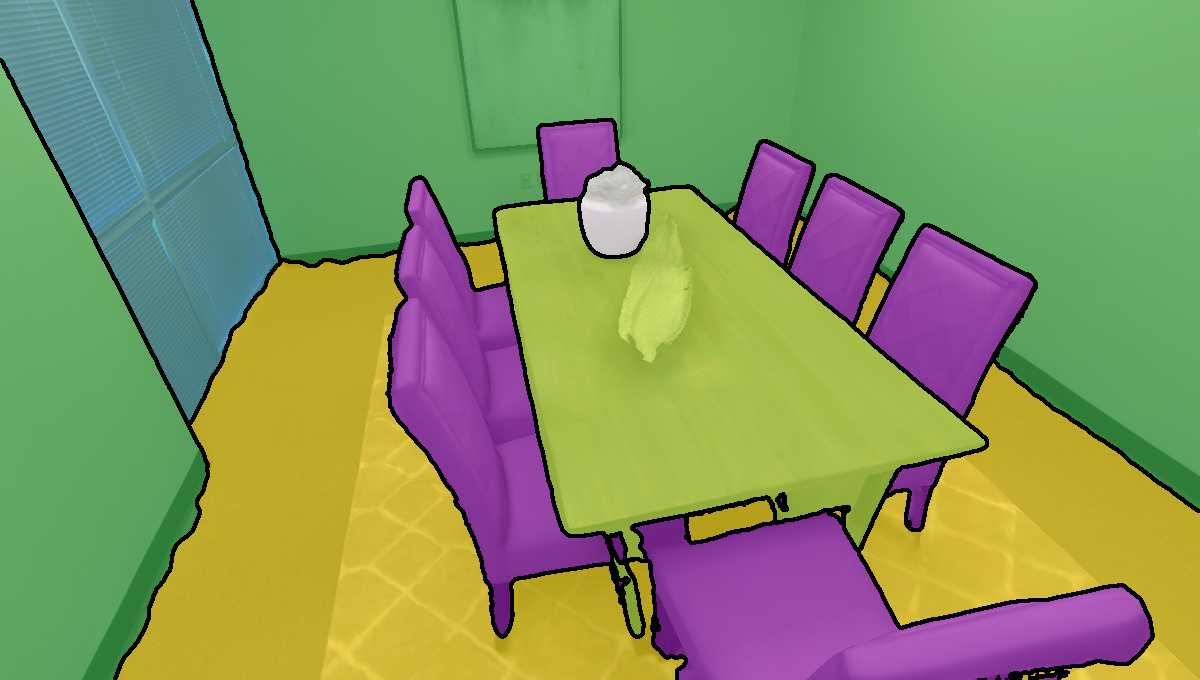}} & 
    \makecell{\includegraphics[height=\sz\columnwidth]{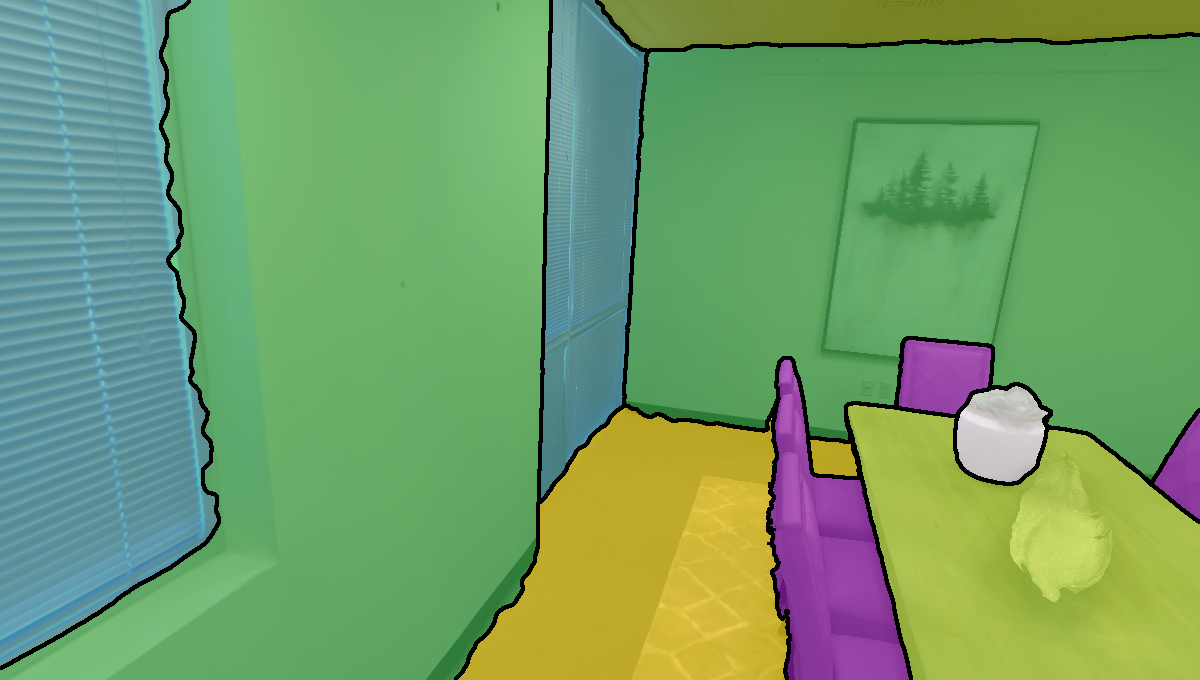}} \\ 
    \end{tabular}
  \caption{Semantic Segmentation Results on Replica~\cite{julian:2019:replica}. We show the multi-view segmentation results of different approaches. 
  The top, middle, and bottom parts show the segmentation results of Mask2Former~\cite{cheng2021mask2former}, our approach without semantic fusion, and our approach with semantic fusion respectively. 
  Comparing the segmentation results from different views, we can see that our method can learn more consistent semantic representation.}
  \label{fig:replica_semantic}
\end{figure*}
\begin{figure}[h]
  \centering
  \scriptsize
  \setlength{\tabcolsep}{1.5pt}
  \newcommand{\sz}{0.31}
  \begin{tabular}{lccc}
    & \tt Room 0 & \tt Room 1 & \tt Room 2 \\
    \makecell{\rotatebox{90}{NICE-SLAM~\cite{zhu:2021:niceslam}}} &
    \makecell{\includegraphics[width=\sz\linewidth]{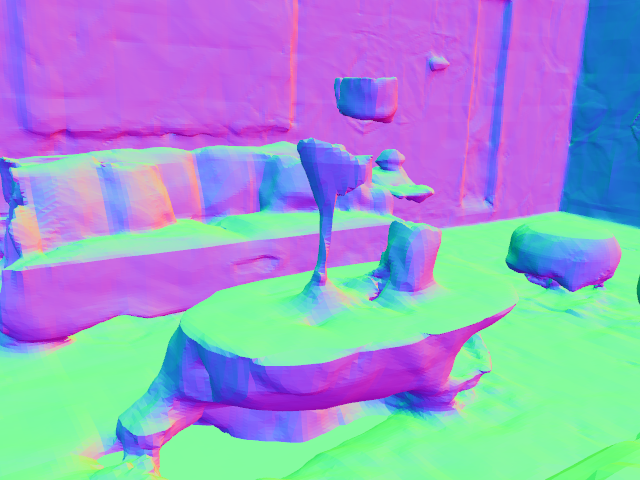}} &
    \makecell{\includegraphics[width=\sz\linewidth]{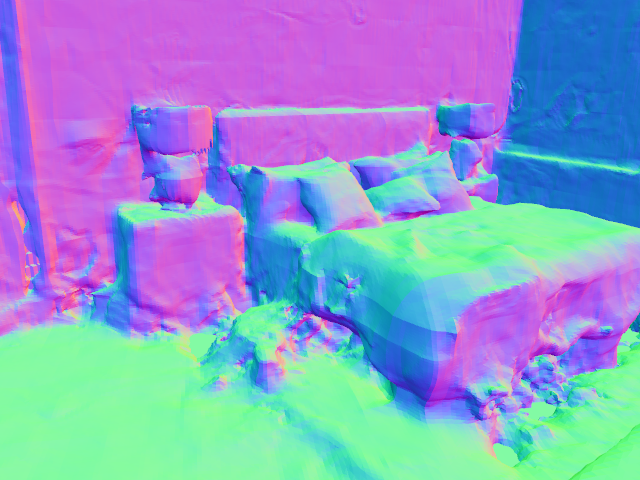}} &
    \makecell{\includegraphics[width=\sz\linewidth]{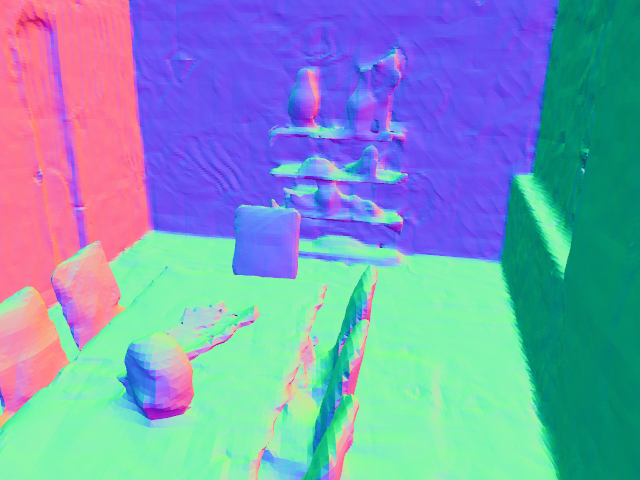}} \\
    \makecell{\rotatebox{90}{Point-SLAM~\cite{point-slam}}} &
    \makecell{\includegraphics[width=\sz\linewidth]{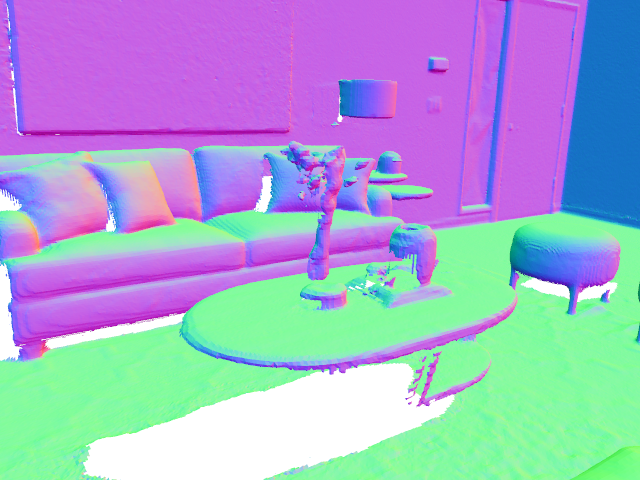}} &
    \makecell{\includegraphics[width=\sz\linewidth]{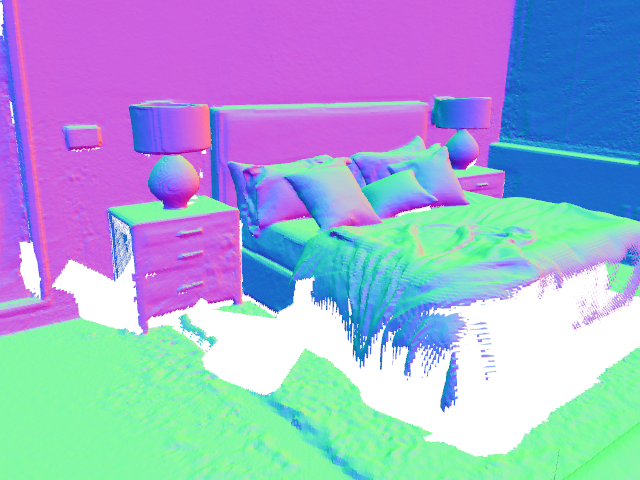}} &
    \makecell{\includegraphics[width=\sz\linewidth]{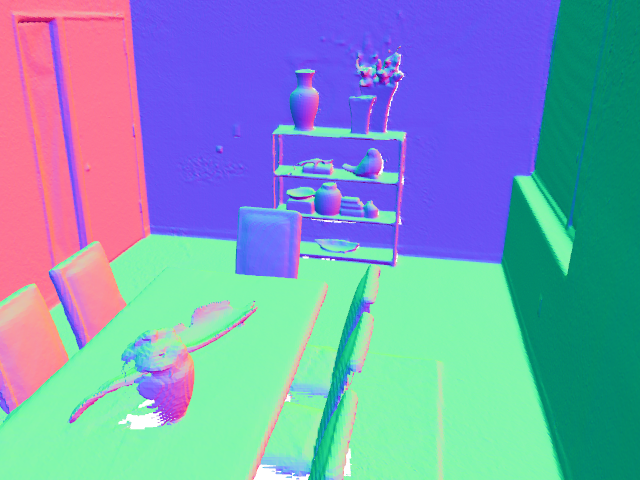}} \\
    \makecell{\rotatebox{90}{Ours}} &    
    \makecell{\includegraphics[width=\sz\linewidth]{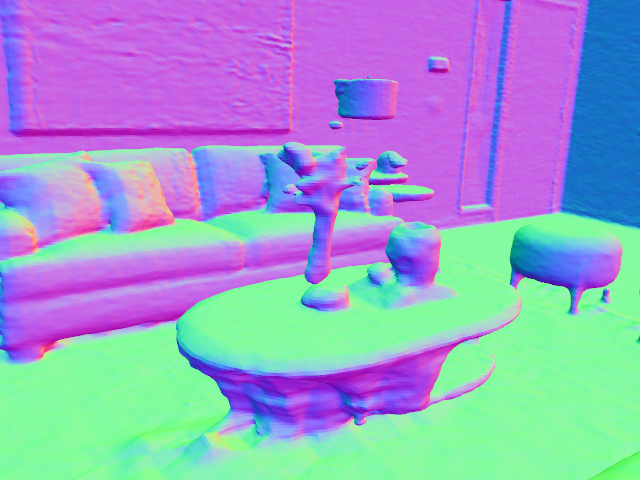}} &
    \makecell{\includegraphics[width=\sz\linewidth]{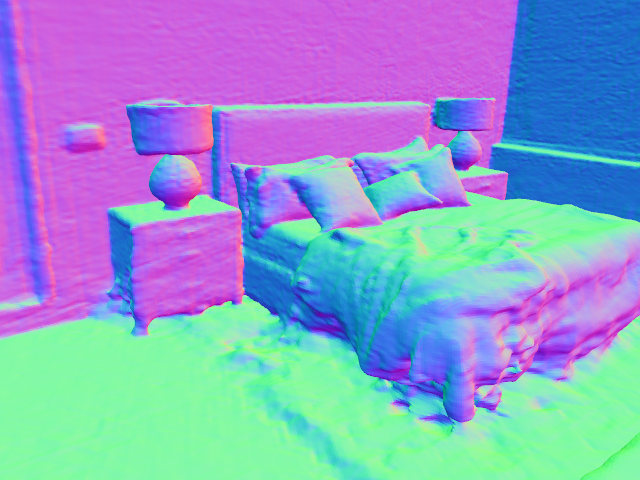}} &
    \makecell{\includegraphics[width=\sz\linewidth]{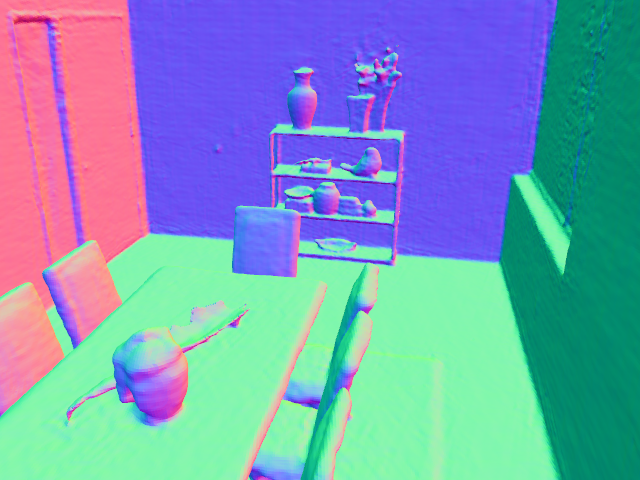}} \\
    \makecell{\rotatebox{90}{GT}} &    
    \makecell{\includegraphics[width=\sz\linewidth]{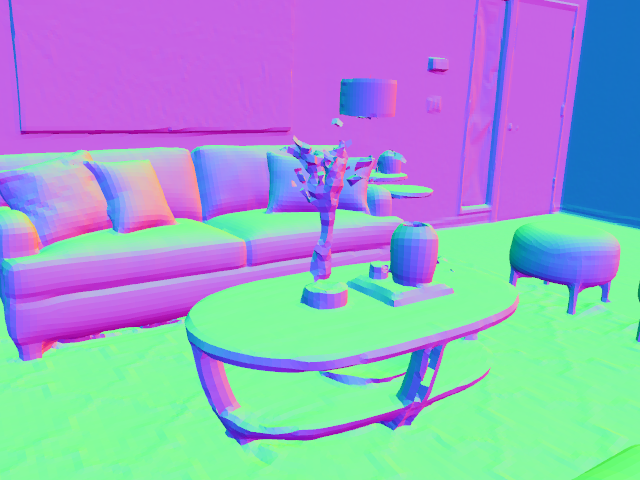}} &
    \makecell{\includegraphics[width=\sz\linewidth]{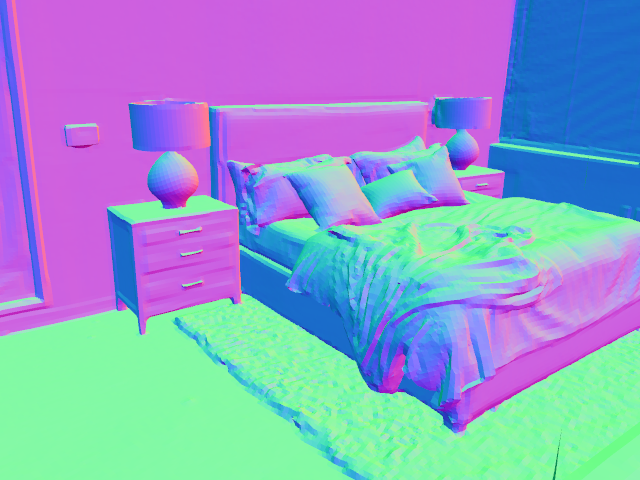}} &
    \makecell{\includegraphics[width=\sz\linewidth]{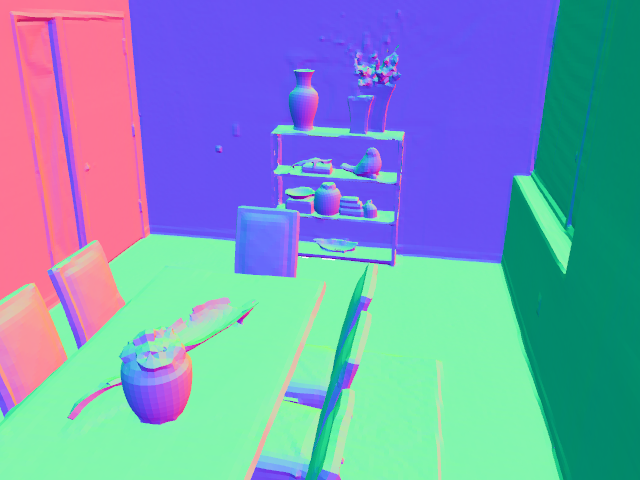}} \\
  \end{tabular}
\caption{Reconstruction Performance on Replica~\cite{julian:2019:replica}. 
We show the novel view rendering results of the reconstructed mesh.}
  \label{fig:replica_mesh}
  \vspace{-4mm}
\end{figure}

\subsection{Evaluation of Scene and Object Reconstruction}
In this part, we evaluate the performance of scene-level and object-level surface reconstruction on Replica dataset~\cite{julian:2019:replica} and ScanNet~\cite{dai:2017:scannet}.
The compared baseline methods, evaluation metrics, and results of this part are shown in the following.

\noindent\textbf{Baseline and Metrics.}
For scene-level reconstruction evaluation, we follow the setting in~\cite{zhu:2021:niceslam,sucar:2021:imap}, we take iMAP~\cite{sucar:2021:imap}, NICE-SLAM~\cite{zhu:2021:niceslam}, Co-SLAM~\cite{co-slam}, and Point-SLAM~\cite{point-slam} as the compared baselines.
The meshes of~\cite{point-slam,co-slam} are generated via their open-source codes.
For object-level reconstruction evaluation, we take vMAP~\cite{kong2023vmap} as the baseline and follow the setting in vMAP.
Due to vMAP needing camera pose as input, we use the camera pose generated via our system as the input of vMAP for a fair comparison.

For the evaluation metrics, we evaluate the generated meshes on \texttt{Accuracy} (cm), \texttt{Completion} (cm), and \texttt{Completion ratio} (\%) with a threshold of 5cm.
Specifically, we sample points $P$ and $Q$ from the reconstructed and ground truth meshes and compute the average distance between a point in one set and its nearest point in the other set.
For scene-level and object-level reconstruction evaluation, we sample 200000 and 10000 points, respectively.

\noindent\textbf{Qualitative and Quantitative Results.}
\cref{tab:recon_replica} shows the quantitative scene-level reconstruction comparison results on the Replica dataset~\cite{julian:2019:replica}.
The results of iMAP and NICE-SLAM are taken from the reported value in~\cite{zhu:2021:niceslam}.
The mesh results of Co-SLAM and Point-SLAM are obtained via their open-source codes. 
Please note that Co-SLAM~\cite{co-slam} manually selected additional virtual camera views to assist mesh culling, and we can't obtain their virtual camera pose.
So, we use the same strategy~\cite{zhu:2021:niceslam} for the meshes generated by Co-SLAM, which leads to better \texttt{Accuracy} (cm) and worse \texttt{Completion} (cm), and \texttt{Completion ratio} (\%) than it reported.
Besides, Point-SLAM~\cite{point-slam} generates mesh using TSDF-Fusion~\cite{dai:2017:bundlefusion}, which leads to a more sparse surface.
So, their \texttt{Accuracy} is relatively better than other approaches, but with worse \texttt{Completion ratio} metrics.

As can be seen from the results in~\cref{tab:recon_replica}, our method achieves better or comparable reconstruction performance, thanks to the accurate camera pose estimation and the inherent smoothness provided by our tetrahedron-based parametric encoding.
We achieved the best results on the \texttt{Completion} (cm), and \texttt{Completion ratio} metrics. 
Compared with the second-best approach, we achieved 6.9\%, and 1.6\% performance improvement on the average results, respectively.
The performance of our approach on \texttt{Accuracy} (cm) metric is only slightly worse than Point-SLAM~\cite{point-slam}, because \cite{point-slam} uses TSDF-Fusion~\cite{dai:2017:bundlefusion} as the post-process to obtain the mesh.
Among those methods whose meshes are generated via marching cubes~\cite{lorensen1998marching}, we achieve the best performance on \texttt{Accuracy} (cm) metric.
Additionally, we show some qualitative comparison results of Replica and ScanNet datasets in the \cref{fig:replica_mesh} and \cref{fig:scannet_mesh}, respectively.
The geometry results in~\cref{fig:replica_mesh} are rendered from novel camera views of three selected Replica scenes.
Our method can restore surface details well and complete the observation area.
The results in~\cref{fig:scannet_mesh} are rendered from the top view to show the overall reconstruction results of the whole scene.
As can be seen, compared with NICE-SLAM~\cite{zhu:2021:niceslam} and Point-SLAM~\cite{point-slam}, we can reconstruct a more watertight, smoother surface for the back side and unobserved areas.

For object-level reconstruction, we compared with vMAP~\cite{kong2023vmap} on the Replica dataset.
Because vMAP~\cite{kong2023vmap} needs the camera poses as the input, we use the camera pose recovered by our system as the input of vMAP for a fair comparison. 
The quantitative comparison results are shown in~\cref{tab:replica_object_recon}.
As can be seen from the table, we achieve the best results on object-level reconstruction.
The decoder parameters of vMAP are approximately $N$ times greater than ours, where $N$ is the number of objects in the scene, as it allocates separate network parameters for each object.
Despite using fewer parameters, we can still estimate the camera pose and achieve comparable reconstruction results.



\begin{table}[htp]
\centering
 \resizebox{1.00\columnwidth}{!}{
  \begin{tabular}{ccccc}
    \toprule
    Method  & \texttt{Acc.}(\%) & \texttt{Avg.Acc.}(\%) & \texttt{mIOU}(\%) & \texttt{fwIOU}(\%)  \\
    \midrule
    Haghighi $^*$~\cite{neural-sem-slam} & 98.47 & 94.14 & 82.61 & 97.22 \\
    Ours $^{*}$ & \textbf{99.09} & \textbf{97.95} & \textbf{98.55} & \textbf{98.20} \\ 
    \hdashline 
    \noalign{\vskip 1pt}
    Mask2Former~\cite{cheng2021mask2former} & 78.84 & 75.66  & 68.20 & 72.75 \\
    Ours w/o Fusion & 82.47 & 80.11 & 69.53 & 75.48\\
    Ours & \textbf{84.88} & \textbf{82.76} & \textbf{70.68} & \textbf{76.72} \\
    \bottomrule
  \end{tabular}
   }
  \caption{Semantic Segmentation Performance on Replica~\cite{julian:2019:replica}. The symbol $*$ denotes the results obtained with ground truth semantic labels. The numbers of~\cite{neural-sem-slam} are taken from their paper.}
  \label{tab:semantic_replica}
\end{table}
\subsection{Evaluation of Semantic Segmentation}
In this part, we evaluate the performance of semantic segmentation on the Replica dataset~\cite{julian:2019:replica}.
The compared baseline methods, evaluation metrics, and results of this part are shown in the following.

\noindent\textbf{Baseline and Metrics.}
For the evaluation of semantic segmentation, we take Haghighi \etal~\cite{neural-sem-slam} as the comparison baseline.
The ground truth semantic labels for Replica~\cite{julian:2019:replica} are rendered with Habitat~\cite{habitat19iccv} with the original trajectory.
Besides, we map the label set following the setting in~\cite{cheng2021mask2former,siddiqui2023panoptic-lifting} for semantic segmentation evaluation.
And following the setting of~\cite{neural-sem-slam}, we use four commonly used metrics: \texttt{Total Accuracy} (\%), \texttt{Avg.Accuracy} (\%), \texttt{mIOU} (\%), and \texttt{fwIOU} (\%).
Besides, we also compare the learned semantic results of our 3D semantic field with the 2D semantic segmentation model Mask2Former~\cite{cheng2021mask2former}. 

\noindent\textbf{Qualitative and Quantitative Results.}
Due to the lack of open-source code for~\cite{neural-sem-slam}, we conducted semantic experiments and comparisons on four scenes (\texttt{Room 0}, \texttt{Room 1}, \texttt{Room 2}, \texttt{Office 0}) based on their settings~\cite{neural-sem-slam}.
The results of average metrics are shown in the top part of~\cref{tab:semantic_replica}.
The symbol \texttt{*} denotes the segmentation results obtained with the supervision of semantic annotation of~\cite{julian:2019:replica}, which is the same to~\cite{neural-sem-slam}.
As can be seen, we outperform~\cite{neural-sem-slam} on all averaged metrics of 4 scenes.
Especially, for the metric $\texttt{mIOU}$, we achieve a significant improvement over~\cite{neural-sem-slam} (82.61$\%$ v.s. 98.55$\%$).

Besides, we also test the performance with 2D noise inputs which are generated by Mask2Former~\cite{cheng2021mask2former} for all scenes.
As shown in the bottom of~\cref{tab:semantic_replica}, compared to the noise 2D segmentation, our approach can improve the multi-view semantic consistency, which leads to 3D consistent semantic representation.
Those quantitative experimental results verify that our approach can learn 3D consistent semantic information from 2D noisy and multi-view inconsistent segmentation results.
Additionally, we also show some qualitative semantic segmentation results in \cref{fig:replica_semantic}.
We show the segmentation results of different views to validate the consistency.
The first row shows different view segmentation results from Mask2Former~\cite{cheng2021mask2former}, the second row shows the results generated by our approaches without semantic fusion (\cref{subsec:sem_fusion}), and the third row shows the results generated by our approaches.
As shown in the figure, the results generated by Mask2Former usually contain multi-view inconsistency, \textit{e.g.} complex edges of furniture, or ambiguous areas.
Under different viewpoints, the same object or region would be recognized as different categories (different colors represent different categories).
Those noisy and inconsistent semantic observations can lead to unstable or conflicting 3D semantic information learning without any process.
For example, the segmentation results in the middle row usually contain a lot of noise and semantic inconsistent areas.
For the performance of our approach (the third row), even with noisy input data, our method can still produce 3D spatially consistent semantic segmentation results across different views.

\subsection{Ablation Study}
\label{subsec:ablation_study}
In this part, we perform the ablation study to investigate the effect of each part in our system.

\begin{table}[h]
\centering
\resizebox{0.9\columnwidth}{!}{
\begin{tabular}{l cc cc}
\toprule
\multirow{2}{*}{Experiment} & \multicolumn{2}{c}{\texttt{Room 1}} & \multicolumn{2}{c}{\texttt{Office 1}} \\
\cmidrule(lr){2-3} \cmidrule(lr){4-5}
 & \textbf{Acc.}[cm]$\downarrow$ &\textbf{Comp.}[cm]$\downarrow$ & \textbf{Acc.}[cm]$\downarrow$ & \textbf{Comp.}[cm]$\downarrow$ \\
\midrule
Only \texttt{PE} &  3.42 & 3.78 & 3.16 & 4.61 \\
+ Cube &  1.30 & 2.48 & 1.14 & 1.88 \\
+ Tetrahedron &  1.25 & 2.37 & 1.09 & 1.67 \\
\bottomrule
\end{tabular}
}
\caption{Reconstruction performance of using different encoding features.}
\label{tab:ablation_representation}
\end{table}
\noindent\textbf{Effect of Multi-view Semantic Fusion.}
The bottom part of \cref{tab:semantic_replica} shows the performance without multi-view semantic fusion, which is lower than our approach.
From the table, we know that without using our strategy, our baseline system can also achieve better semantic segmentation results than Mask2Former~\cite{cheng2021mask2former}. 
This is because our neural implicit representation can learn 3D consistent representation from noise 2D input, and progressively converge to a global optimal from 2D multi-view observations.

\noindent\textbf{Effect of Using Different Neural Representation.}
Here, we show the reconstruction performance of using different neural representations for our system.
We select two scenes from Replica~\cite{julian:2019:replica} for evaluation and the results are shown in~\cref{tab:ablation_representation}.
As can be seen from the table, without any additional feature encoding, only using \texttt{PE} can lead to a worse reconstruction of the scene.
Using additional features can enhance the encoding capacity of MLPs and lead to performance improvement.
Besides, we also show the performance of using additional multi-resolution cube-based feature encoding~\cite{mueller2022instantngp}.
Compared with the cube representation, our tetrahedron representation can achieve a better surface reconstruction.

\begin{figure}[ht]
\centering
\includegraphics[width=0.49\linewidth]{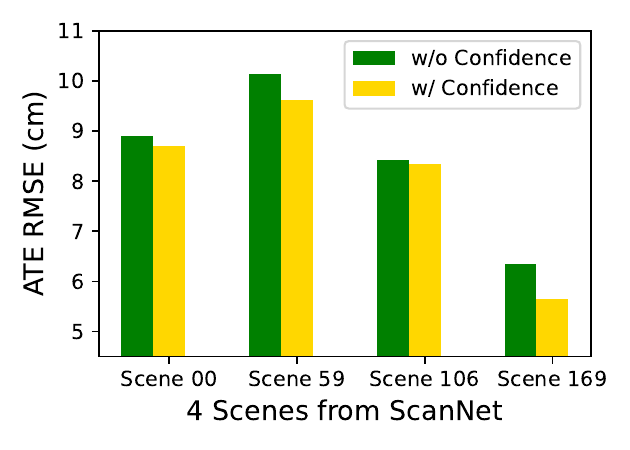}
\includegraphics[width=0.49\linewidth]{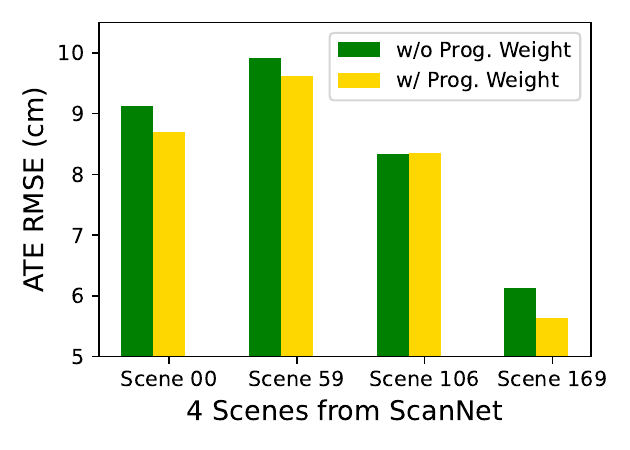}
 \caption{Tracking Performance on 4 Scenes from ScanNet~\cite{dai:2017:scannet} (ATE RMSE $\downarrow$ [cm]). The left and right figures respectively show the effects of using confidence-based pixel sampling and progressive optimization weight during the camera tracking process.}
\label{fig:ablation_confidence_sample}
\end{figure}

\noindent\textbf{Effect of Confidence-guided Pixel Sampling.}
Here, we show the results to validate the effectiveness of confidence-based pixel sampling for camera tracking.
We select four scenes from ScanNet~\cite{dai:2017:scannet} dataset for evaluation and the results are shown in the left part of~\cref{fig:ablation_confidence_sample}.
As can be seen from the figure, with the guidance of semantic confidence we can reduce the camera tracking error (\texttt{ATE RMSE}) on the selected scenes.
With the help of confidence, we can avoid sampling pixels in the object edges areas and complex geometric regions during the tracking process.
The pixels in these areas usually have a relatively slow convergence speed in the mapping process, which will lead to relatively large tracking errors.
 
\noindent\textbf{Effect of Progressive Optimization Weight.}
We show the performance of using progressive optimization weight for camera tracking.
The comparison results of 4 selected scenes are shown in the right part of~\cref{fig:ablation_confidence_sample}.
As can be seen from the figure, with progressive optimization weight we can also reduce the camera tracking error (\texttt{ATE RMSE}) on the selected scenes.
The progressive weight can lead to a coarse-to-fine optimization manner and avoid the model falling into local minima.

\subsection{Memory and Running Time}
\label{subsec:time}
\begin{figure}[ht!]
\centering
\includegraphics[width=0.9\linewidth]{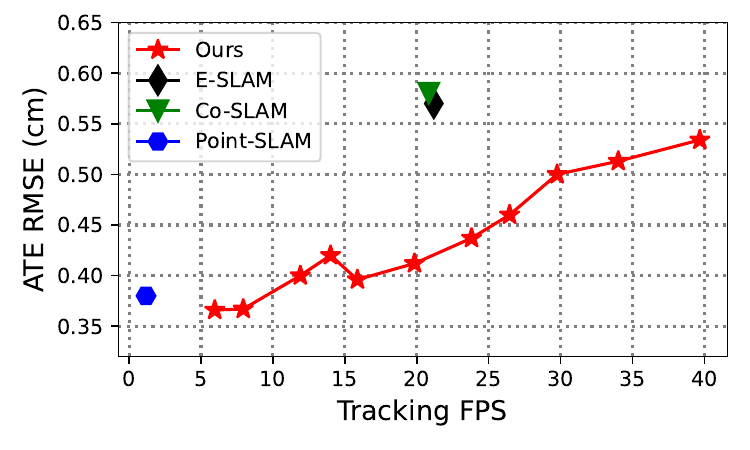}
 \caption{Tracking speed (FPS) and error (ATE RMSE) of different approaches. 
We use different tracking iterations to obtain the ATE RMSE erorr curve at different tracking FPS.}
\label{fig:ablation_fps}
\end{figure}

In this part, we show the memory usage and runtime performance of our approach on the Replica dataset.

For the memory usage of our system, the whole training parameters of our system are three MLP decoders \{$\mathcal{M}_{geo}$,$\mathcal{M}_{rgb}$, $\mathcal{M}_{sem}$\} and multi-resolution tetrahedron features $\Theta$.
For example, on scene \texttt{Office 0}, the whole parameters of MLP decoders and embedding features are 8.02MB (0.02MB + 8.00MB), which is much less than the parameters of Point-SLAM, 27.74MB (0.51MB + 27.23MB).
For the running speed of our system, we show the camera tracking FPS with respect to the \texttt{ATE RMSE} (cm) on the Replica dataset.
We perform global bundle adjustments for 10 iterations when a new keyframe comes, and use 5-50 iterations for camera tracking.
The tracking FPS results with different tracking iterations are shown in~\cref{fig:ablation_fps}.
The runtimes are profiled on a single NVidia RTX 4090 card. 
We can achieve good tracking performance at the \textasciitilde30 tracking FPS with about 0.5 cm tracking error.
Additionally, we also show the performance of ESLAM~\cite{eslam}, Co-SLAM~\cite{co-slam}, and Point-SLAM~\cite{point-slam} in the figure.
Compared to ESLAM and Co-SLAM, we can achieve better camera tracking accuracy at the same speed.
Compared to Point-SLAM, we can achieve faster tracking speed with comparable tracking performance.

\begin{figure}[ht!]
\centering
\includegraphics[width=0.99\linewidth]{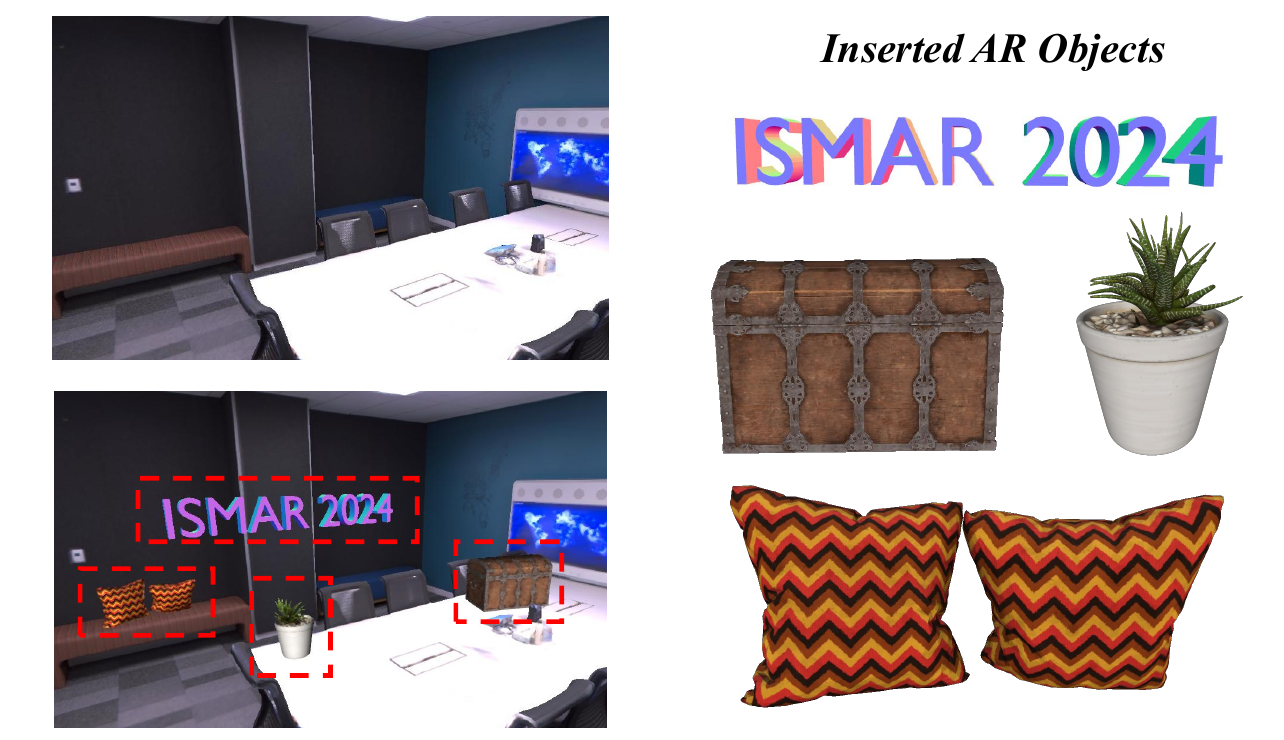}
 \caption{The AR demo on scene \texttt{Office 4} from Replica~\cite{julian:2019:replica}. The top-left is the original scene image, and on the right are three AR objects and the logo of \texttt{ISMAR 2024}. 
 We render those objects in the original scene with the camera pose estimated by our approach. 
 The rendered results are shown in the lower left.
 For more AR demo videos, please refer to our supplementary material.}
\label{fig:ar_demo}
\end{figure}

\subsection{AR Application}
\label{subsec:application}
In this part, we show the application of our system in AR.
Our system can not only localize the pose of the camera but also reconstruct watertight scene geometry/depth and perform novel view synthesis via our implicit representation. 
Therefore, our method can be very suitable for VR/AR applications. 
We demonstrate an AR example, as shown in~\cref{fig:ar_demo}.
The scene image is selected from \texttt{Office 4} of Replica~\cite{julian:2019:replica} dataset.
We place three virtual objects and one \texttt{ISMAR 2024} logo (shown on the right of the figure) into reconstructed scenes according to the pose and depth recovered by our system.
The rendered AR image is shown at the bottom of~\cref{fig:ar_demo}, and we highlighted the virtual objects and logo with red boxes.
We can accurately represent the occlusion relationship between real and virtual contents via our accurate camera pose and rendered dense depth map.
For the multi-view AR demo, please refer to our supplementary video, which shows we can handle occlusion between different objects very well.

\section{Conclusion}
\label{sec:conclusion}
In this paper, we present NIS-SLAM, a neural implicit dense semantic RGB-D SLAM that can model consistent semantic information from inconsistent segmentation results generated by a pre-trained 2D CNN.
For high-fidelity surface reconstruction and spatial
consistent scene understanding, we use a hybrid representation of high-frequency multi-resolution tetrahedron features and low-frequency positional encoding for our system.
And the multi-view semantic fusion is proposed to handle the inconsistency of 2D segmentation results.
Besides, the semantic guided pixel sampling and progressive optimization weight are used for robust camera tracking.
Extensive experiments on a variety of datasets have shown the effectiveness and application of our proposed system. 

Currently, the proposed NIS-SLAM approach relies on the segmentation results of a close-set model, which limits its applications for open-set world tasks.
Combined with the large language/open-set models may be more practical for many applications.
Compared to methods with explicit representation (point, 3D Gaussian), our approach cannot recover high-frequency information well.

\section{ACKNOWLEDGMENTS}
This work was partially supported by NSF of China  (No.61932003).

\newpage
\bibliographystyle{abbrv-doi-hyperref}
\bibliography{draft}

\end{document}